\documentclass{article}
\usepackage[main,final]{neurips_2026}
\usepackage[utf8]{inputenc} 
\usepackage[T1]{fontenc}    
\usepackage{hyperref}       
\hypersetup{
    hidelinks
}
\usepackage{url}            
\usepackage{algorithmic}
\usepackage{tikz}
\usetikzlibrary{arrows.meta, positioning}
\usepackage{booktabs}       
\usepackage{xcolor}
\definecolor{darkgreen}{rgb}{0,0.5,0}  
\usepackage{amsfonts}       
\usepackage{amsmath}
\usepackage{nicefrac}       
\usepackage{booktabs}
\usepackage{multirow}
\usepackage{amssymb}
\usepackage{enumitem}
\usepackage{makecell}
\usepackage{siunitx}   
\usepackage{microtype}      
\usepackage{xcolor}         
\usepackage{graphicx}      
\title{Cyclic Discrete Diffusion: A Novel Multi-Class Segmentation Refinement Technique}

\author{%
  Libo Chen \\
  Department of Mathematical Sciences \\
  University of Bath \\
  Bath, United Kingdom \\
  \texttt{lc2762@bath.ac.uk}
  \And
  Souvik Ghosh \\
  International Institute of Information Technology \\
  Hyderabad, India \\
  \texttt{souvik.g@research.iiit.ac.in}
  \And
  Vinay P.~Namboodiri \\
  Department of Computer Science \\
  University of Bath \\
  Bath, United Kingdom \\
  \texttt{vpn22@bath.ac.uk}
}

\begin{document}
\maketitle

\begin{abstract}
	We cast image segmentation refinement as a \textit{discrete} diffusion process. To solve this task we introduce Cyclic Discrete Diffusion (CDD), a novel forward–reverse formulation with a cyclic label topology that maps a discrete label set to the uniform distribution and back. Instead of injecting Gaussian noise as is done for continuous diffusion, CDD evolves segmentation masks through label jumps: at each step, every pixel either remains in its current class or jumps to the next label with a fixed rate. This jump-or-stay mechanism defines a finite-state Markov process with bounded forward and reverse steps, featuring analytically consistent and tractable reverse dynamics. 
    CDD operates natively in a multi-class label space, enabling joint 
    refinement of all classes in a single pass without requiring class-wise decomposition, as commonly needed in prior binary refinement approaches. The proposed formulation admits a finite and controllable number of refinement steps, 
    for which we analyze the convergence properties through spectral gap analysis. We evaluate CDD across diverse segmentation tasks, including multi-class semantic, 
    instance-level, and binary segmentation refinement. Our method consistently improves both region accuracy and boundary quality across multiple backbones and datasets.
\end{abstract}

\section{Introduction}

\label{sec:intro}
Segmentation is a fundamental task in computer vision, providing pixel-level understanding of visual scenes and object structures~\cite{long2015fully, ronneberger2015u, chen2017deeplab}.
Despite substantial advances~\cite{he2017mask, ke2022mask, kirillov2020pointrend}, generating precise and coherent segmentation masks remains a challenge, especially in maintaining boundary precision and structural consistency across regions.
The difficulty is particularly acute in high-resolution and multi-class scenes, where complex boundaries, overlapping semantics, and subtle visual transitions must be handled simultaneously. While general methods like Mask-RCNN are widely applicable and used, we argue that they are not precise enough, especially at high resolutions \cite{peng2022pp, yuan2020segfix}. This observation has led to a number of methods that are focused on refinement of segmentation \cite{yuan2020segfix, peng2022pp, wu2025medical, wang2023segrefiner}. A number of these methods are specific to particular architectures or modalities- for instance BPR \cite{peng2022pp} is specific to Mask-RCNN while \cite{wu2025medical} is specific to the medical domain. This limits their generalization and extensibility. 

\begin{figure}
    \centering
    \includegraphics[width=0.91\textwidth]{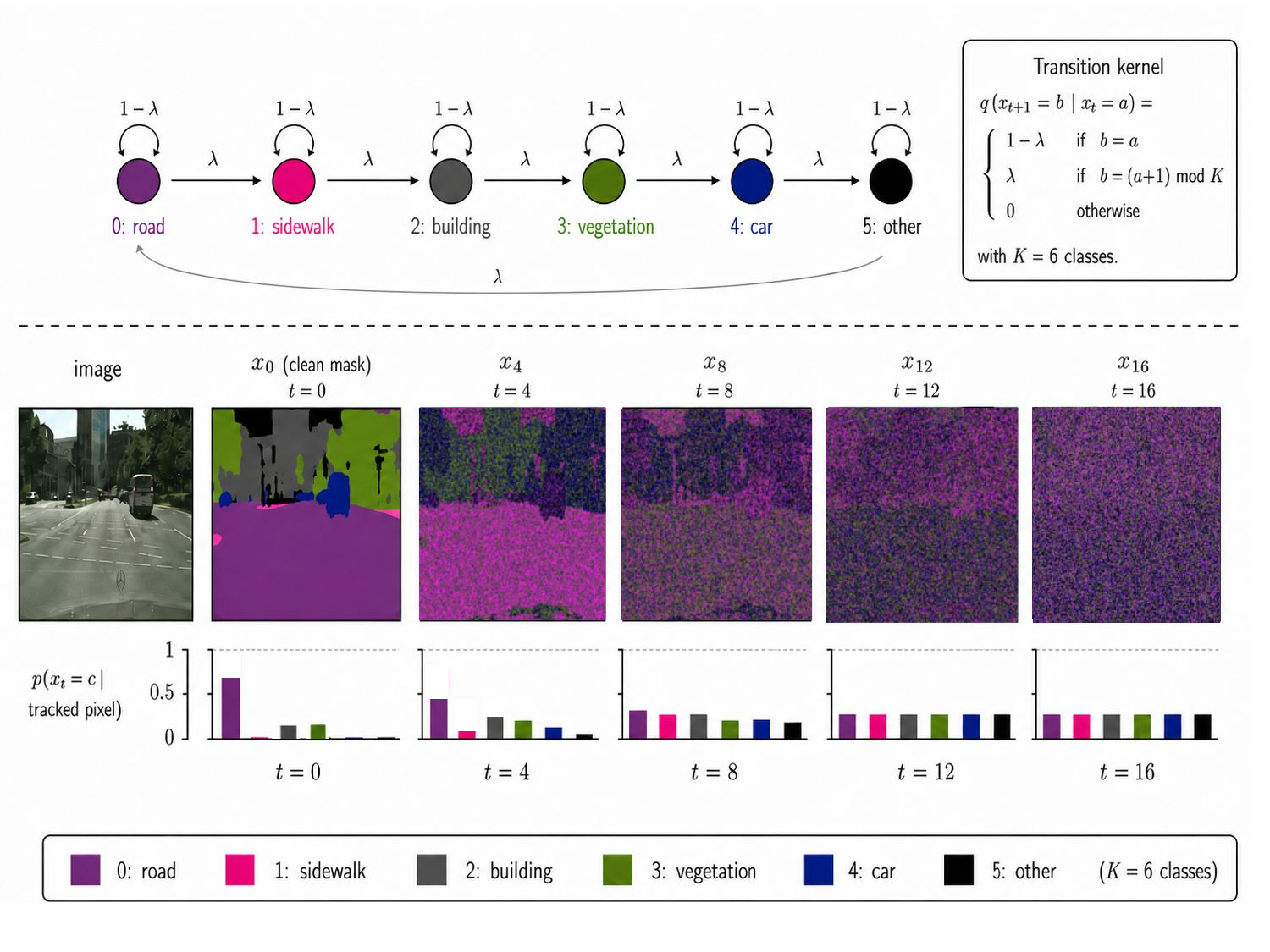}
\caption{Method illustration of the proposed cyclic discrete diffusion (CDD).
Each pixel label evolves under the cyclic transition kernel $q(x_{t+1}\!\mid\!x_t)$: it stays in the current class with probability $1-\lambda$ or jumps to the next class (mod $K$) with probability $\lambda$.
Starting from a clean mask ($x_0$), the forward process progressively corrupts the segmentation over time ($t=0,4,8,12,16$), driving per-pixel class probabilities to an approximately uniform mixture (shown by the bars).
Our model learns a reverse transition $p_\theta(x_{t-1}\!\mid\!x_t)$ to recover a refined mask.}
    \label{fig:CDD_idea}
\end{figure}

An aspect that has been recently studied in this context is model-agnostic segmentation refinement ~\cite{yuan2020segfix,shen2022high, wang2023segrefiner}.
This task enables us to refine the outputs of any pretrained segmentation network, making it widely applicable and effective in practice.
Among these, SegRefiner~\cite{wang2023segrefiner}, is closest to our work. They show that this problem can be solved using a conditional discrete diffusion framework, enabling gradual refinement of coarse predictions into fine-grained masks.\\
While this approach makes an interesting contribution, we identify that it could result in practical challenges for usage. For instance, typically one uses multi-class semantic segmentation. In order to use SegRefiner as designed, one has to do \textit{binary segmentation refinement} by taking each class into account and refining the mask. This poses a significant practical challenge. To address this challenge, we pose a novel discrete diffusion based \textit{multi-class segmentation refinement} approach that solves the problem in a single pass.  

\begin{figure*}
    \centering        
    \includegraphics[width=0.878\textwidth]{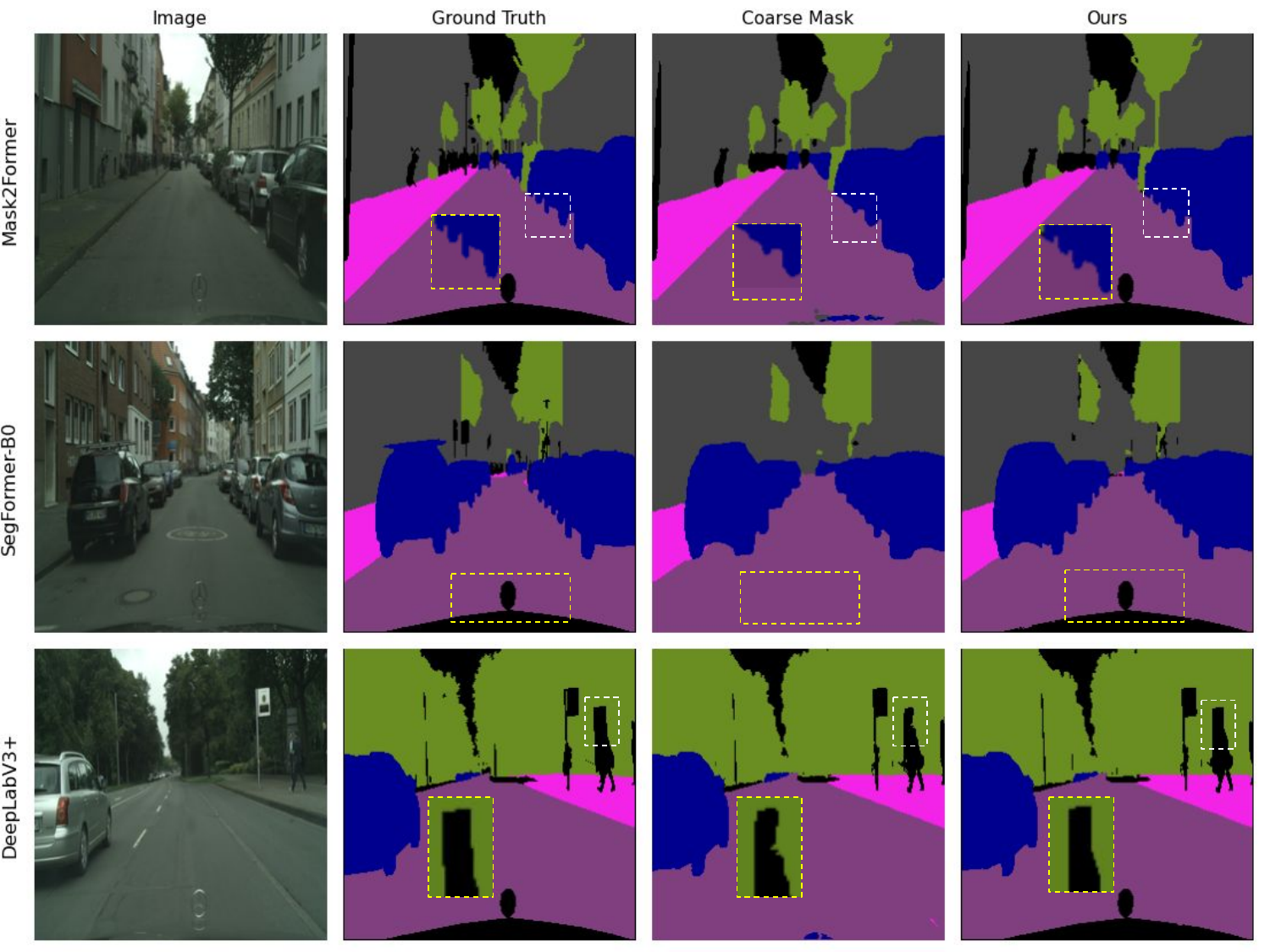}
    \caption{
    Qualitative comparison of multi-class segmentation refinement on Cityscapes.
    For each backbone, we show the input image, ground truth, coarse mask, and our CDD-refined result.
    CDD consistently improves coarse predictions from different segmentation models, yielding cleaner boundaries and better local structural consistency.
    Highlighted regions show representative boundary and fine-structure corrections.
    }
    \label{fig:dis3_results}
\end{figure*}


Our design follows a Markov jump process termed a cyclic discrete diffusion (CDD). To enable us to solve for multi-class refinement, we follow this jump process over the label space. In our formulation, each pixel follows a jump rule at every time step, either jumping to the next discrete state with a fixed probability or remaining in its current state. This process is illustrated in fig. \ref{fig:CDD_idea}. To ensure stability in discrete space, we further define a cyclic state connection, allowing transitions to wrap around the label space. This cyclic construction guarantees that the process converges to a uniform stationary distribution, establishing a mathematically consistent and well-posed Markov chain. Intuitively, the jump-or-stay mechanism captures the stochastic evolution of segmentation boundaries, where each pixel probabilistically decides whether to update its class assignment, aligning closely with the iterative refinement nature of segmentation.
Second, CDD provides an analytically tractable forward process with a finite number of explicitly controllable steps.
We further derive a theoretical analysis to calculate the number of refinement steps, offering principled guidance for step selection and ensuring efficient convergence. Together, these properties form a principled bridge between discrete probabilistic modeling and real-world segmentation refinement.

We extensively evaluate CDD across diverse segmentation tasks and backbones. 
We first evaluate CDD on multi-class semantic segmentation benchmarks, where 
it performs joint refinement across all classes in a single pass, leading to 
substantial improvements in both IoU and boundary quality. These results 
highlight the advantage of modeling segmentation refinement directly in a 
multi-class discrete space.
To enable direct comparison with prior work such as SegRefiner, which is 
designed for binary mask refinement, we additionally evaluate CDD in the 
binary setting on standard benchmarks. On the COCO validation set with LVIS 
annotations \cite{lin2014microsoft, gupta2019lvis}, CDD consistently improves 
over strong refinement baselines and outperforms SegRefiner across multiple 
backbones in both Mask AP and Boundary AP 
(Tab.~\ref{tab:coco_val_comparison}, Tab.~\ref{tab2:instance_segmentation_models}). 
These results confirm the effectiveness, generality, and strong boundary precision 
of our principled multi-class discrete diffusion framework for segmentation refinement.


\textbf{Contributions:} This work makes the following novel contributions:
\begin{itemize}
\item We generalize the idea of discrete diffusion to multi-class refinement in a single pass providing significant practical benefits using a principled framework of jumps in a Markov chain.
\item We propose a novel transition matrix thereby extending the framework of D3PM \cite{austin2021structured}. Our transition probabilities are derived based on the Markov Jump Process.  We provide extensive results in Section~\ref{sec:experiments} that demonstrate the efficacy of our method.
\item We also provide a theoretical analysis that guides the estimation of the number of refinement steps and provides evidence of convergence based on the spectral gap analysis.
\end{itemize}

\section{Related Work}
\subsection{Segmentation Refinement}
Segmentation refinement aims to enhance coarse or low-quality masks by improving boundary precision and recovering fine-grained structures.
Early methods often used probabilistic graphical models, such as fully connected CRFs~\cite{krahenbuhl2011efficient}, to enforce local smoothness and align region boundaries with image edges.
Deep networks like FCNs~\cite{long2015fully} later integrated refinement into end-to-end learning, though traditional CRF-based post-processing operates independently of deep features and has limited adaptability to complex visual patterns.
With the rise of deep learning, refinement has evolved into a learnable and task-specific process.
Representative works, such as the Deep Watershed Transform~\cite{bai2017deep} and Boundary-Aware Instance Segmentation~\cite{hayder2017boundary}, train networks to predict truncated distance or boundary maps within cropped instances, achieving more precise delineation of object contours.
PointRend~\cite{kirillov2020pointrend} formulated segmentation as adaptive point-based rendering, focusing computational effort on uncertain regions. 
There are some model-agnostic approaches~\cite{tang2021look, zhou2020deepstrip, cheng2020cascadepsp,shen2022high}. While these methods achieve strong empirical gains and remain efficient in practice, they are generally designed around specific segmentation pipelines and rely on deterministic updates, limiting their adaptability to multi-class or highly ambiguous regions.  
Several more general methods have been proposed.
For instance,  SegFix~\cite{yuan2020segfix}  proposed a model-agnostic boundary correction framework that explicitly predicts pixel-wise displacement vectors to repair misaligned edges.
More recently, SegRefiner~\cite{wang2023segrefiner} unified mask-level refinement across diverse instance segmentation models, progressively enhancing boundary details through an iterative UNet-like architecture. We show that our method outperforms these more recent general works using a principled approach.

\subsection{Diffusion Models in Continuous and Discrete Spaces}
Continuous diffusion models have achieved remarkable success in image synthesis and other continuous domains by parameterizing the gradient of the data distribution through score matching~\cite{song2019generative,ho2020denoising,song2020score}. 
These models define a forward process that gradually perturbs data with Gaussian noise and a reverse process learned to denoise samples step by step, resulting in high-quality and diverse generations. 
However, their reliance on continuous noise makes them inherently ill-suited for discrete structures such as segmentation labels, where transitions between categorical states cannot be modeled by infinitesimal Gaussian perturbations.
To address this issue, discrete diffusion models~\cite{austin2021structured,campbell2022continuous,sun2022score,lou2023discrete} reformulate the forward–reverse dynamics as a Markov process, introducing categorical corruption and discrete transition matrices. 
Although these methods extend the diffusion paradigm to token- or label-based data, they often suffer from unstable training objectives and limited scalability. 
For example, mean prediction formulations~\cite{austin2021structured} and ratio-matching losses~\cite{sun2022score} provide theoretical consistency but yield suboptimal empirical performance, while more recent score-entropy formulations~\cite{lou2023discrete} improve stability at the cost of additional computational complexity. We further validate that, for segmentation refinement, Segrefiner~\cite{wang2023segrefiner} that follows~\cite{austin2021structured} indeed demonstrates suboptimal performance.
RefineCatDiff~\cite{liu2025refinecatdiff} also employs
categorical diffusion for medical segmentation refinement.
Unlike general categorical transitions, CDD uses a cyclic
jump-or-stay kernel with only two admissible predecessor
states, yielding a tractable reverse posterior.
Our work builds upon this line of research by introducing a CDD that defines a finite, cyclic, and interpretable transition structure in discrete space.
Unlike prior discrete diffusion models, CDD incorporates a jump-or-stay stochastic rule governed by a controllable rate parameter, ensuring both convergence to a uniform stationary distribution and coherent coarse-to-fine refinement.

\subsection{Preliminaries: Diffusion Models (Continuous and Discrete)}
\label{sec:method:diffusion}

In continuous diffusion model, the data are gradually transformed into noise through a forward stochastic differential equation (SDE)~\cite{song2020score}. 
$x_0$ denote a data sample drawn from the data distribution, and $x_t$ its perturbed version at time~$t$. 
The forward SDE is defined as
\begin{equation} \label{fsde}
    d x_t = f(x_t, t)\, dt + g(t)\, dw_t,
\end{equation}
where $f$ and $g$ are smooth drift and diffusion coefficients, and $w_t$ denotes standard Brownian motion. 
As $t \to T$, the distribution of $x_t$ converges to a prior noise distribution, typically Gaussian.

The goal of the reverse process is to invert this stochastic dynamics and recover the original data distribution from noise. 
The corresponding reverse SDE is given by ~\cite{anderson1982reverse}
\begin{equation} \label{sde}
    d x_t = \big(f(x_t, t) - g^2(t)\, \nabla \log p(x_t, t)\big)\, dt + g(t)\, d \overline{w}_t,
\end{equation}
where $\nabla \log p(x_t, t)$ is the score function, i.e., the gradient of the log-density at time~$t$, and $\overline{w}_t$ is a Brownian motion adapted to the reverse-time filtration.

To extend diffusion modeling to categorical data, ~\cite{austin2021structured} proposed the Structured Denoising Diffusion Model (D3PM), 
which generalizes continuous diffusion to finite discrete state spaces. 
Instead of adding Gaussian noise, the forward process applies random state transitions governed by a Markov transition matrix $Q_t \in \mathbb{R}^{K\times K}$:
\begin{equation}
    q(x_t \mid x_{t-1}) = \mathrm{Cat}(x_t; p = x_{t-1} Q_t),
    \label{eq:d3pm-forward}
\end{equation}
where $\mathrm{Cat}(x_t; p)$ denotes a \textit{categorical distribution} over $K$ discrete states with class probabilities, $x_t \in \{1,2,\dots,K\}$ represents a categorical variable at time~$t$. 
This process progressively corrupts the original data by repeatedly sampling from $Q_t$, 
driving the distribution of $x_t$ toward a predefined stationary distribution.

The reverse process is trained to approximate the true posterior $q(x_{t-1}\mid x_t, x_0)$, 
which follows directly from the Markov chain and can be expressed analytically via Bayes’ rule.
A neural network $f_\theta(x_t, t)$ predicts the reverse transition probabilities 
$p_\theta(x_{t-1}\mid x_t)$, and the model is optimized via a variational lower bound 
equivalent to a per-step cross-entropy objective.

\section{Our Method: Cyclic Discrete Diffusion}
In this work, we propose CDD, a novel discrete diffusion–based framework inspired by the dynamics of cyclic (clock-like) jumps.
Unlike continuous Gaussian diffusion models that operate in Euclidean space by adding Gaussian perturbations, CDD evolves over a finite discrete label space through stochastic state transitions.
Each pixel either jumps to its neighboring class or remains unchanged, according to a fixed transition probability, forming a cyclic Markov process over semantic labels.
This process defines a discrete-time Markov chain that progressively mixes the categorical states and converges to a uniform terminal distribution.
In the forward process, CDD gradually perturbs the ground-truth mask into a highly randomized mask by repeatedly applying these local jumps.
In the reverse process, the model starts from a noisy mask conditioned on a coarse input and reconstructs the clean segmentation by learning the reverse transition probabilities.
The following sections detail the forward and reverse formulations and present the overall learning objective.

\subsection{Discrete Forward Process}
The forward process in CDD progressively perturbs the ground-truth segmentation mask into a noisy one through a sequence of discrete stochastic jumps.
Let $\mathcal{X}=\{0,1,\dots,K-1\}$ denote the finite set of semantic labels, and let $x_t \in \mathcal{X}$ be the class of a pixel at timestep~$t$.
At each step, the pixel either \emph{jumps} to its next class or \emph{remains} in the current state. Formally, the transition kernel is defined as
\begin{equation}
  q(x_{t+1}=b \mid x_t=a) =
  \begin{cases}
    1-\lambda, & b=a, \\
    \lambda, & b=(a+1) \bmod K, \\
    0, & \text{otherwise}
  \end{cases}
  \label{eq:forward}
\end{equation}
where $\lambda \in (0,1)$ is the jump rate.
The use of the modulo operator “$\bmod K$” plays a key role in this design. 
It makes the label transitions cyclic: 
when a pixel in the last class $K-1$ performs a jump,
it wraps around to the first class $0$ instead of
stopping at the boundary.
Unlike an open random walk that terminates at the edge, this closed-loop construction ensures that all labels are mutually reachable.
This forward design leads to a Markov chain that can not only traverse the entire label space, but also converge to a uniform stationary distribution.
Let $u$ denote the stationary probability of being in label~$x$ after an infinite number of jumps.
Then, for the proposed cyclic process, all labels share the same probability
\begin{equation}
  u(x) = \tfrac{1}{K}, \quad \forall x \in \mathcal{X}.
  \label{eq:uniform}
\end{equation}
The circular design is particularly suited for segmentation tasks, which provides both theoretical tractability and practical suitability for segmentation refinement.
Unlike Gaussian perturbations in continuous space, which can blur class boundaries, 
our discrete cyclic dynamics operate directly on categorical labels, 
preserving semantic consistency while enabling controllable degeneracy.

As $t$ increases, the structured mask is gradually randomized through many jumps,
eventually converging to a uniform categorical distribution. Let $\pi_t$ be the current marginal distribution at time $t$.
Formally, the total variation distance between $\pi_t$ and the stationary distribution decays exponentially as
\begin{equation}
\|\pi_t - u\|_{\mathrm{TV}} \le C e^{-\gamma t},
\end{equation}
where the effective exponential decay rate of the cyclic transition kernel, for \(0<\lambda<1\), is
\begin{equation}
\gamma =
-\frac{1}{2}\log\left[
1-2\lambda(1-\lambda)
\left(1-\cos\frac{2\pi}{K}\right)
\right].
\end{equation}

The decay rate $\gamma$ depends only on the jump rate $\lambda$ and the number of classes $K$.
This closed-form expression guarantees fast mixing and ensures that the process
reaches near-uniformity within a small number of iterations,
approximately $t_{\mathrm{mix}}(\epsilon) \approx \tfrac{1}{\gamma}\log(1/\epsilon)$
(see Appendix~\ref{subsec:mix_rate} for derivation and empirical validation).

\subsection{Discrete Reverse Process}

The goal of the reverse process is to invert the forward jump dynamics defined in~\eqref{eq:forward} 
and recover a structured segmentation mask from noise. 
The forward transition in~\eqref{eq:forward} follows a ``jump-or-stay'' rule, 
Formally, if a pixel is in state~$x_t$ at step~$t$,
it can only come from two possible previous states: $x_t$ (no jump) or $x_t-1$ (one cyclic jump).
\footnote{All label operations are computed modulo~$K$, consistent with the forward process.}
This discreteness makes the reverse transition analytically tractable. Let $\pi_{t-1}(x \mid Y)$ denote the marginal distribution of the forward process at step~$t\!-\!1$ for label~$x$, 
optionally conditioned on auxiliary information~$Y$ (e.g., coarse images).
Intuitively, $\pi_{t-1}(x \mid Y)$ quantifies the likelihood that a pixel takes label~$x$ just before the next jump in the forward direction.

For a given target label~$x_t$, 
the total incoming probability mass to~$x_t$ from step~$t\!-\!1$ is
\begin{equation}
D_t(x_t \mid Y)
= \lambda\,\pi_{t-1}(x_t{-}1 \mid Y)
+ (1-\lambda)\,\pi_{t-1}(x_t \mid Y),
\label{eq:post-denom}
\end{equation}
where the two terms respectively correspond to (i) arriving at~$x_t$ via a jump from~$x_t\!-\!1$ with rate~$\lambda$, 
and (ii) remaining in~$x_t$ with probability~$1-\lambda$.

By Bayes' rule, the reverse transition kernel can then be expressed in closed form as the posterior over these two admissible predecessors:
\begin{equation}
\begin{aligned}
q(x_{t-1}=x_t{-}1 \mid x_t, Y)
= \frac{\lambda\,\pi_{t-1}(x_t{-}1 \mid Y)}{D_t(x_t\mid Y)},
\quad q(x_{t-1}=x_t \mid x_t, Y)
= \frac{(1-\lambda)\,\pi_{t-1}(x_t \mid Y)}{D_t(x_t\mid Y)}.
\end{aligned}
\label{eq:posterior-compact}
\end{equation}
Equation~\eqref{eq:posterior-compact} gives the exact reverse transition probabilities 
of the forward jump process in~\eqref{eq:forward}.
It specifies, for each pixel and timestep~$t$, 
how likely we are to ``undo'' the last jump or to ``stay'' when stepping backward from~$x_t$ to~$x_{t-1}$.

\subsection{Training Objectives}

Given the image $I$, coarse mask $Y$, noisy mask $x_t$, and timestep $t$, the network predicts per-pixel categorical probabilities $p_\theta(x_0\mid x_t,I,Y)$. 
The RGB image, one-hot coarse mask, and one-hot noisy
mask are concatenated along the channel dimension,
while the coarse mask remains fixed throughout
reverse sampling.
Following~\cite{wang2023segrefiner}, we use a reconstruction loss and a texture-preserving regularization, and further add a CDD posterior matching term. The overall objective is
\begin{equation}\label{eq:loss}
\mathcal{L}
=
\alpha_{\mathrm{CE}}\mathcal{L}_{\mathrm{CE\mbox{-}X0}}
+
\alpha_{\mathrm{tex}}\mathcal{L}_{\mathrm{texture}}
+
\alpha_{\mathrm{KL}}\mathcal{L}_{\mathrm{KL}} .
\end{equation}

The reconstruction loss is a weighted cross-entropy:
\begin{equation}
\mathcal{L}_{\mathrm{CE\mbox{-}X0}}
=
-\mathbb{E}_{q(x_0,t,x_t)}
\left[
\frac{1}{|\Omega|}\sum_{u\in\Omega}
w(u)\log p_\theta(x_0(u)\mid x_t,I,Y)
\right],
\end{equation}
where $w(u)=4$ if $Y(u)\neq x_0(u)$ and $w(u)=1$ otherwise.

The texture loss is computed class-wise between the softmax probabilities $P_\theta$ and the one-hot ground-truth mask $G$:
\begin{equation}
\mathcal{L}_{\mathrm{texture}}
=
\left\|S(P_\theta)-S(G)\right\|,
\end{equation}
where $S(\cdot)$ denotes Sobel gradient magnitudes after $3\times3$ mean-filter smoothing.

To enforce consistency with the analytical CDD reverse dynamics, we use a posterior matching loss. Since one reverse step only allows $x_{t-1}\in\{x_t,(x_t-1)\bmod K\}$, the true posterior $q(x_{t-1}\mid x_t,x_0)$ is a two-point distribution. Following the $x_0$-parameterization of discrete diffusion models~\cite{austin2021structured}, we induce the reverse posterior by marginalizing over the predicted clean label:
\begin{equation}
p_\theta(x_{t-1}\mid x_t,I,Y)
\propto
\sum_{\tilde{x}_0=0}^{K-1}
q(x_{t-1}\mid x_t,\tilde{x}_0)\,
p_\theta(\tilde{x}_0\mid x_t,I,Y),
\end{equation}
and minimize
\begin{equation}
\mathcal{L}_{\mathrm{KL}}
=
\mathbb{E}_{x_0\sim p_{\mathrm{data}},\ t\sim \mathrm{Unif}(\{1,\dots,T\}),\ x_t\sim q(\cdot\mid x_0,t)}
\left[
D_{\mathrm{KL}}
\left(
q(x_{t-1}\mid x_t,x_0)
\,\|\, 
p_\theta(x_{t-1}\mid x_t,I,Y)
\right)
\right].
\end{equation}

We set $\alpha_{\mathrm{CE}}=1$, $\alpha_{\mathrm{KL}}=10^{-3}$, and $\alpha_{\mathrm{tex}}=5$. During training, half of the samples use $x_t$ generated by the CDD forward process from $x_0$, while the other half use $x_t=Y$ and $t=T$, so that the model directly observes the real coarse-mask distribution used at inference time. 
The ablation in Table~\ref{tab:mixing_ratio}
shows that the 50/50 training strategy maintains
strong performance under both single-pass and
iterative inference.

\vspace{-1.5mm}
\section{Segmentation Refinement Experiments}
\label{sec:experiments}

\subsection{Multi-Class Semantic Segmentation Refinement}

We evaluate multi-class segmentation refinement on the Cityscapes dataset~\cite{cordts2016cityscapes}. Cityscapes is a standard benchmark for urban-scene semantic segmentation, containing complex street scenes with multiple interacting object categories and fine boundary structures. 
The training split is used for model optimization, and all results are reported on the validation split. 
In our multi-class refinement setting, we map the original semantic labels into six categories: road, sidewalk, building, vegetation, car, and background/others. 
This six-class setting allows us to test whether CDD can handle multi-class label refinement, while avoiding the excessive class imbalance and optimization difficulty of the full Cityscapes label space.
Ground-truth semantic masks serve as the refinement targets, and coarse masks are produced by different segmentation backbones, including Mask2Former~\cite{cheng2022masked}, SegFormer-B0~\cite{xie2021segformer}, and DeepLabV3+~\cite{chen2018encoder}. 
All masks are resized to $256 \times 256$ for both training and evaluation.

For CDD, we set the number of classes to $K=6$, the number of diffusion steps to $T=16$(see Appendix~\ref{subsec:mix_rate}) , and the jump rate to $\lambda=0.3$. 
The model is trained with the objective in Eq.~\eqref{eq:loss}, using Adam with a learning rate of $10^{-4}$ and a batch size of $8$. 
Training is run for at most $120$ epochs with early stopping, where the patience is set to $15$ epochs and the relative improvement threshold on the training loss is $10^{-3}$. 
We report Intersection-over-Union (IoU)~\cite{everingham2010pascal} for region-level segmentation quality and BFScore~\cite{perazzi2016benchmark} for boundary accuracy.

\subsubsection{Quantitative Results}

Table~\ref{tab:main_cityscapes_refinement} evaluates whether CDD can consistently improve coarse masks from different segmentation backbones.
Averaged over Mask2Former, SegFormer-B0, and DeepLabV3+, our method improves IoU by $11.04$ points and BFScore by $17.09$ points. 
Across all three backbones, the gain in BFScore is larger than the gain 
in IoU. This suggests that the improvement is not only due to correcting 
large semantic regions, but also comes from better boundary alignment. 
This behavior can be explained by the loss design. 
The texture-preserving loss explicitly constrains spatial gradients of the predicted mask, 
encouraging sharper transitions at object boundaries, 
while the weighted CE loss emphasizes regions where the coarse mask disagrees with the ground truth, 
which are typically concentrated around boundaries. Figure~\ref{fig:dis3_results} provides qualitative visualization results.
We further evaluate zero-shot refinement on stronger
backbones, SegNeXt-L~\cite{guo2022segnext} and SegMAN-L~\cite{fu2025segman}, using CDD trained
only on Mask2Former coarse masks. As shown in
Table~\ref{tab:main_cityscapes_refinement},
CDD consistently improves both IoU and BFScore,
demonstrating its transferability to unseen backbones.

Table~\ref{tab:multiclass_refinement_comparison} compares CDD with representative refinement baselines on the Cityscapes six-class setting, using Mask2Former coarse masks as input. 
To assess the contribution of diffusion-based refinement,
we include a non-diffusion CNN post-processing baseline
that directly refines the coarse mask using the RGB image.
As shown in Table~\ref{tab:multiclass_refinement_comparison},
CDD outperforms this baseline by 6.80 IoU and 7.06 BFScore
points under single-pass inference, demonstrating its
effectiveness beyond conventional CNN-based refinement.
D3PM~\cite{austin2021structured} serves as a generic discrete diffusion baseline for multi-class refinement. 
In its standard form, D3PM performs conditional refinement by starting from a noisy state sampled from the terminal distribution and conditioning on the coarse mask during reverse sampling, but it yields only marginal improvements over the coarse baseline. 
SegRefiner~\cite{wang2023segrefiner} is originally designed for binary mask refinement, so we adapt it to our six-class setting by training one model per class and merging the class-wise predictions into a multi-class mask. Although SegRefiner improves over the coarse mask, its gains are smaller than those of CDD, suggesting that binary refinement strategies do not directly transfer optimally to multi-class semantic refinement.
CDD obtains the best IoU and BFScore among all compared methods. 
We further analyze the two discrete diffusion processes in the Appendix~\ref{subsec:multi_class}, where CDD shows clearer advantages in multi-step reverse sampling from uniform noisy masks, with improved reverse-process stability and boundary preservation.

\begin{table}[t]
\centering
\caption{Multi-class segmentation refinement on Cityscapes.
We report IoU and BFScore (\%) across different backbones.
For zero-shot evaluation, CDD is trained only on
Mask2Former coarse masks without retraining.}
\label{tab:main_cityscapes_refinement}
\setlength{\tabcolsep}{5pt}
\begin{tabular}{lcccccc}
\toprule
\multirow{2}{*}{\textbf{Coarse model}}
& \multicolumn{3}{c}{\textbf{IoU (\%)}} 
& \multicolumn{3}{c}{\textbf{BFScore (\%)}} \\
\cmidrule(lr){2-4} \cmidrule(lr){5-7}
& Coarse & Ours & $\Delta$
& Coarse & Ours & $\Delta$ \\
\midrule

\multicolumn{7}{l}{\textit{Original evaluation backbones}} \\

Mask2Former
& 73.00 & \textbf{84.55} & +11.55
& 63.46 & \textbf{82.20} & +18.74 \\

SegFormer-B0
& 71.17 & \textbf{82.47} & +11.30
& 60.44 & \textbf{78.93} & +18.49 \\

DeepLabV3+
& 75.41 & \textbf{85.69} & +10.28
& 70.32 & \textbf{84.35} & +14.03 \\

\cmidrule(lr){1-7}
Average (above three)
& 73.19 & 84.24 & +11.04
& 64.74 & 81.83 & +17.09 \\

\midrule
\multicolumn{7}{l}{\textit{Zero-shot transfer (unseen backbones)}} \\

SegNeXt-L
& 77.27 & \textbf{85.09} & +7.82
& 76.90 & \textbf{85.42} & +8.52 \\

SegMAN-L
& 78.04 & \textbf{85.47} & +7.43
& 76.11 & \textbf{85.07} & +8.96 \\

\bottomrule
\end{tabular}
\end{table}

\begin{table}[t]
\centering
\setlength{\tabcolsep}{5pt}
\caption{Comparison of multi-class segmentation
refinement methods on Cityscapes using Mask2Former
coarse masks. We report IoU and BFScore (\%) and
their improvements over the coarse baseline.}
\label{tab:multiclass_refinement_comparison}
\begin{tabular}{lcccc}
\toprule
Method & IoU & $\Delta$IoU & BFScore & $\Delta$BF \\
\midrule
Coarse mask
& 73.00 & -- & 63.46 & -- \\

CNN post-processing
& 77.75 & +4.75 & 75.14 & +11.68 \\

D3PM
& 72.80 & -0.20 & 64.56 & +1.10 \\

SegRefiner
& 78.48 & +5.48 & 71.35 & +7.89 \\

CDD (Ours)
& \textbf{84.55} & \textbf{+11.55}
& \textbf{82.20} & \textbf{+18.74} \\
\bottomrule
\end{tabular}
\end{table}

\subsection{Binary Mask Refinement}
Although our main contribution is native multi-class segmentation refinement, 
the existing diffusion-based segmentation refinement method~\cite{wang2023segrefiner} is designed and evaluated in the binary mask setting. 
Following SegRefiner~\cite{wang2023segrefiner}, 
we evaluate our method on three representative benchmarks that cover both 
instance-level and pixel-level segmentation scenarios:
COCO~\cite{lin2014microsoft}, 
LVIS~\cite{gupta2019lvis}, 
and DIS5K~\cite{qin2022highly}(see appendix(\ref{subsec:dis_result_section})). 
COCO provides large-scale and diverse everyday scenes for general instance segmentation, 
while LVIS focuses on fine-grained, large-vocabulary segmentation with challenging boundary details. 
DIS5K contains high-resolution images with delicate object contours and 
serves as a strong benchmark for evaluating fine-grained segmentation refinement. 
For all datasets, we follow the same experimental protocol as SegRefiner~\cite{wang2023segrefiner}, including the official training and validation splits, coarse predictions, training losses, and evaluation metrics, to ensure a fair comparison.
We evaluate model performance using IoU, mBA(mean
Boundary Accuracy)~\cite{cheng2020cascadepsp}, Mask Average Precision (Mask AP), and Boundary Average Precision (Boundary AP) \cite{cheng2021boundary}.
Mask AP measures region-level accuracy based on the IoU overlap between predicted and ground-truth masks, whereas Boundary AP focuses on contour alignment within a narrow band around object edges, making it particularly sensitive to fine-grained boundary details and small structures.
\subsubsection{Quantitative Results}

\begin{table*}[htbp]
\centering
\caption{Comparison with refinement baselines on the COCO val set using LVIS annotations.}
\label{tab:coco_val_comparison}
\scriptsize
\setlength{\tabcolsep}{2.5pt}
\begin{tabular}{lcccccc@{\hspace{6pt}}cccccc}
\toprule
\multirow{2}{*}{\textbf{Method}} &
\multicolumn{6}{c}{\textbf{Mask AP}} &
\multicolumn{6}{c}{\textbf{Boundary AP}} \\
\cmidrule(lr){2-7} \cmidrule(lr){8-13}
& AP & AP$_{50}$ & AP$_{75}$ & AP$_S$ & AP$_M$ & AP$_L$
& AP & AP$_{50}$ & AP$_{75}$ & AP$_S$ & AP$_M$ & AP$_L$ \\
\midrule
Mask R-CNN-R50
& 39.8 & 61.4 & 42.3 & 24.9 & 47.0 & 55.1
& 27.3 & 53.3 & 25.2 & 24.9 & 41.8 & 27.4 \\
+ SegFix~\cite{yuan2020segfix}
& 40.6 & 61.4 & 42.8 & 25.0 & 48.4 & 56.6
& 29.1 & 53.7 & 28.0 & 24.9 & 43.6 & 30.8 \\
+ BPR~\cite{tang2021look}
& 41.0 & 61.4 & 43.1 & 24.8 & 48.5 & 57.8
& 30.4 & 55.2 & 29.5 & 24.7 & 43.8 & 33.7 \\
+ SegRefiner~\cite{wang2023segrefiner}
& 41.9 & 61.6 & 43.2 & 25.7 & 49.4 & 58.8
& 32.6 & 55.7 & \textbf{32.5} & 25.6 & 45.0 & 37.3 \\
\textbf{+ Ours}
& \textbf{42.4} & \textbf{61.9} & \textbf{44.3} & \textbf{25.9} & \textbf{50.1} & \textbf{59.1}
& \textbf{32.6} & \textbf{56.2} & 32.2 & \textbf{25.8} & \textbf{45.7} & \textbf{37.5} \\
\midrule
Mask R-CNN-R101
& 41.6 & 63.3 & 44.4 & 26.5 & 49.5 & 57.8
& 29.0 & 55.2 & 26.7 & 26.3 & 44.4 & 29.8 \\
+ SegFix~\cite{yuan2020segfix}
& 42.2 & 63.4 & 44.7 & 26.5 & 50.9 & 59.1
& 30.6 & 56.1 & 30.0 & 26.3 & 46.0 & 33.1 \\
+ BPR~\cite{tang2021look}
& 42.8 & 63.3 & 45.3 & 26.1 & 51.0 & 60.6
& 32.0 & 57.3 & 31.6 & 25.9 & 46.3 & 36.3 \\
+ SegRefiner~\cite{wang2023segrefiner}
& 43.6 & 63.3 & 45.2 & 27.4 & 51.4 & 61.6
& 34.1 & 57.2 & 34.9 & 27.2 & 47.1 & \textbf{39.9} \\
\textbf{+ Ours}
& \textbf{44.2} & \textbf{63.7} & \textbf{46.7} & \textbf{27.4} & \textbf{52.5} & \textbf{61.6}
& \textbf{34.3} & \textbf{58.3} & \textbf{35.0} & \textbf{27.2} & \textbf{48.3} & 39.3 \\
\bottomrule
\end{tabular}
\end{table*}

\begin{table*}[t]
\centering
\caption{Comparison of baseline models and our Cyclic Discrete Diffusion refinement (“+ Ours”) on the LVIS val set (\%). green numbers indicate improvements over the base model.}
\label{tab2:instance_segmentation_models}
\resizebox{0.96\linewidth}{!}{
\begin{tabular}{l|llll|llll}
\toprule
\multirow{2}{*}{\textbf{Method}} 
& \multicolumn{4}{c|}{\textbf{Mask AP (\%)}} 
& \multicolumn{4}{c}{\textbf{Boundary AP (\%)}} \\
\cmidrule(lr){2-5}\cmidrule(lr){6-9}
& AP & AP$_S$ & AP$_M$ & AP$_L$ 
& AP & AP$_S$ & AP$_M$ & AP$_L$ \\
\midrule

\textbf{CondInst}~\cite{tian2020conditional}          
& 39.8 & 24.3 & 47.5 & 54.8
& 29.2 & 24.1 & 42.6 & 30.2 \\
+ SegRefiner 
& 41.1 & 25.5 & 48.8 & 56.8 
& 32.2 & 25.3 & 44.5 & 36.1 \\
+ Ours 
& \textbf{41.8} {\color{darkgreen}(+2.0)} & \textbf{26.2} {\color{darkgreen}(+1.9)} & \textbf{50.0} {\color{darkgreen}(+2.5)} & \textbf{56.4} {\color{darkgreen}(+1.6)} 
& \textbf{33.2} {\color{darkgreen}(+4.0)} & \textbf{24.6} {\color{darkgreen}(+0.5)} & \textbf{45.4} {\color{darkgreen}(+2.8)} & \textbf{39.6} {\color{darkgreen}(+9.4)} \\

\midrule
\textbf{SOLO}~\cite{wang2020solo}            
& 37.4 & 19.3 & 45.5 & 56.6
& 24.7 & 19.0 & 39.3 & 27.8 \\
+ SegRefiner 
& 40.5 & 21.7 & 49.3 & 60.9 
& 31.3 & 21.4 & 44.8 & 39.8 \\
+ Ours 
& \textbf{41.0} {\color{darkgreen}(+3.6)} & \textbf{22.4} {\color{darkgreen}(+3.1)} & \textbf{49.7} {\color{darkgreen}(+4.2)} & \textbf{60.2} {\color{darkgreen}(+3.6)} 
& \textbf{31.2} {\color{darkgreen}(+6.5)} & \textbf{22.2} {\color{darkgreen}(+3.2)} & \textbf{45.2} {\color{darkgreen}(+5.9)} & \textbf{37.9} {\color{darkgreen}(+10.1)} \\

\midrule
\textbf{PointRend}~\cite{kirillov2020pointrend}
& 41.5 & 25.1 & 49.0 & 59.3
& 30.6 & 25.0 & 44.2 & 34.1 \\
+ SegRefiner 
& 42.8 & 25.9 & 50.4 & 61.3 
& 33.7 & 25.8 & 46.2 & 40.1 \\
+ Ours 
& \textbf{43.0} {\color{darkgreen}(+1.5)} & \textbf{25.7} {\color{darkgreen}(+0.6)} & \textbf{50.8} {\color{darkgreen}(+1.8)} & \textbf{60.9} {\color{darkgreen}(+1.6)} 
& \textbf{33.7} {\color{darkgreen}(+3.1)} & \textbf{25.7} {\color{darkgreen}(+0.7)} & \textbf{46.5} {\color{darkgreen}(+2.3)} & \textbf{39.2} {\color{darkgreen}(+5.1)} \\

\midrule
\textbf{QueryInst}~\cite{fang2021instances}
& 42.4 & 26.7 & 49.5 & 61.4
& 29.9 & 26.4 & 44.5 & 32.3 \\
+ SegRefiner
& 44.2 & 27.5 & 51.4 & 64.7
& 34.9 & 27.3 & 47.2 & 41.6 \\
+ Ours
& \textbf{45.1} {\color{darkgreen}(+2.7)} & \textbf{27.9} {\color{darkgreen}(+1.2)} & \textbf{52.2} {\color{darkgreen}(+2.7)} & \textbf{65.0} {\color{darkgreen}(+3.6)}
& \textbf{35.5} {\color{darkgreen}(+5.6)} & \textbf{28.1} {\color{darkgreen}(+1.7)} & \textbf{48.1} {\color{darkgreen}(+3.6)} & \textbf{41.7} {\color{darkgreen}(+9.4)} \\

\midrule
\textbf{Mask Transfiner}~\cite{ke2022mask}
& 43.2 & 25.9 & 49.0 & 60.1
& 31.6 & 25.8 & 44.5 & 35.8 \\
+ SegRefiner 
& 43.3 & 26.8 & 49.9 & 62.0 
& 34.4 & 26.6 & 45.9 & 41.3 \\
+ Ours 
& \textbf{44.5} {\color{darkgreen}(+1.3)} & \textbf{27.9} {\color{darkgreen}(+2.0)} & \textbf{52.3} {\color{darkgreen}(+3.3)} & \textbf{63.0} {\color{darkgreen}(+2.9)} 
& \textbf{35.1} {\color{darkgreen}(+3.5)} & \textbf{27.8} {\color{darkgreen}(+2.0)} & \textbf{48.1} {\color{darkgreen}(+3.6)} & \textbf{41.3} {\color{darkgreen}(+5.5)} \\

\midrule
\textbf{Mask2Former}~\cite{cheng2022masked}
& 46.8 & 27.9 & 54.9 & 69.1
& 37.0 & 27.8 & 50.1 & 44.6 \\
+ SegRefiner 
& 47.4 & 28.5 & 55.3 & 69.8 
& 38.8 & 28.4 & 51.0 & 48.5 \\
+ Ours 
& \textbf{48.1} {\color{darkgreen}(+1.3)} & \textbf{29.0} {\color{darkgreen}(+1.1)} & \textbf{56.3} {\color{darkgreen}(+1.4)} & \textbf{70.1} {\color{darkgreen}(+1.0)} 
& \textbf{39.1} {\color{darkgreen}(+2.1)} & \textbf{28.7} {\color{darkgreen}(+0.9)} & \textbf{51.9} {\color{darkgreen}(+1.8)} & \textbf{48.9} {\color{darkgreen}(+4.3)} \\

\midrule
\textbf{RefineMask}~\cite{wang2020solo}  
& 41.2 & 24.0 & 48.1 & 59.2
& 30.5 & 23.8 & 43.5 & 34.1 \\
+ SegRefiner 
& 41.9 & 24.6 & 48.7 & 60.7 
& 33.0 & 24.5 & 44.7 & 39.4 \\
+ Ours 
& \textbf{42.4} {\color{darkgreen}(+1.2)} & \textbf{24.6} {\color{darkgreen}(+0.6)} & \textbf{49.6} {\color{darkgreen}(+1.5)} & \textbf{61.0} {\color{darkgreen}(+1.8)} 
& \textbf{33.2} {\color{darkgreen}(+2.7)} & \textbf{24.6} {\color{darkgreen}(+0.8)} & \textbf{45.4} {\color{darkgreen}(+1.9)} & \textbf{39.6} {\color{darkgreen}(+5.5)} \\

\bottomrule
\end{tabular}}
\label{tab2:lvis_full_group_compare}
\end{table*}

\vspace{-2.25mm}
\paragraph{Instance-level Refinement on COCO}
We evaluate CDD on the COCO validation set with LVIS annotations.
For comparison, we select the three representative model-agnostic refinement baselines:
SegFix~\cite{yuan2020segfix}, BPR~\cite{tang2021look}, and SegRefiner~\cite{wang2023segrefiner}.
These methods are designed to refine the coarse masks produced by segmentation models
without retraining the underlying detectors, forming a fair basis 
for evaluating the refinement quality of our CDD. Following their setup, the coarse masks used in Table~\ref{tab:coco_val_comparison} 
are obtained from Mask R-CNN to ensure consistency with the original experiments 
in~\cite{tang2021look,yuan2020segfix}. 
As shown in Table~\ref{tab:coco_val_comparison}, CDD achieves the strongest overall performance across both Mask R-CNN backbones. 
Compared with SegRefiner, which already provides a strong refinement baseline, CDD further improves most Mask AP and Boundary AP metrics, indicating that the improvements are consistent across both region-level mask quality and boundary alignment.

\vspace{-2.25mm}
\paragraph{Cross-model Generalization on LVIS}
To further evaluate transferability, we apply CDD as a plug-and-play refinement module to 7 representative instance segmentation models.
As shown in Table~\ref{tab2:instance_segmentation_models}, our method consistently improves performance across all baselines.
For example, CDD achieves a gain of up to +3.6 AP and +6.5 Boundary AP on SOLO, 
and significantly boosts all model's Boundary AP\textsubscript{L} 
by +9.4 on CondInst, demonstrating its strong capability in refining large-object boundaries.
Compared with SegRefiner, our approach maintains superior results across all models, 
confirming that the proposed cyclic discrete diffusion mechanism 
can be effectively integrated into diverse segmentation frameworks. 
The underlying jump-and-stay mechanism enables more accurate recovery of fine boundary details,
leading to higher segmentation precision and sharper mask contours.
Qualitative comparisons in Fig.~\ref{fig:coco_lvis} further illustrate these improvements,
where CDD refines complex shapes and restores more consistent object boundaries than existing refinement methods.

\vspace{-2.5mm}
\section{Conclusion and Discussion}
\label{sec:conclusion}
We presented Cyclic Discrete Diffusion (CDD), a principled discrete diffusion framework for segmentation refinement. Unlike prior binary refinement approaches, CDD operates directly in a multi-class label space and performs joint refinement of all classes in a single pass. The proposed cyclic jump process defines a structured Markov transition over discrete labels, yielding a mathematically tractable forward process with a finite number of controllable refinement steps and a uniform distribution. We further derived the corresponding reverse dynamics and provided a spectral-gap-based analysis of convergence.
Experiments across semantic, instance-level, and binary refinement benchmarks demonstrate that CDD consistently improves both region accuracy and boundary quality across diverse segmentation backbones. Comparisons against existing discrete diffusion formulations further suggest that the structured cyclic transition provides a suitable inductive bias for segmentation refinement by preserving partial semantic structure during corruption and reverse denoising.
While our primary focus is segmentation refinement, the proposed framework also opens directions for broader discrete generative modeling in structured label spaces. Future work may explore scaling the framework to larger semantic datasets. 

\paragraph{Computational costs and limitations.}
CDD requires only 1.15 GPU-hours for training
(120 epochs, NVIDIA H200), compared with 1.20
GPU-hours for a matched feed-forward U-Net.
Single-pass inference takes 16.5 ms per image
on an RTX A5000, compared with 12.1 s for
SegRefiner under its full refinement protocol
on the same GPU. These results demonstrate
the computational efficiency of CDD, although
its scalability to higher resolutions and
larger label spaces remains to be investigated.

\begin{ack}
\paragraph{Funding.}
Libo Chen acknowledges support from the University of
Bath through the Vice-Chancellor's International
Scholarship. Computational resources for this work
were provided by the Department of Computer Science
at the University of Bath. No dedicated project
funding was received for this research.

\paragraph{Competing Interests.}
The authors declare no competing interests.
\end{ack}

\bibliographystyle{plainnat}
\bibliography{main}

\begin{thebibliography}{41}
\providecommand{\natexlab}[1]{#1}
\providecommand{\url}[1]{\texttt{#1}}
\expandafter\ifx\csname urlstyle\endcsname\relax
  \providecommand{\doi}[1]{doi: #1}\else
  \providecommand{\doi}{doi: \begingroup \urlstyle{rm}\Url}\fi

\bibitem[Anderson(1982)]{anderson1982reverse}
Brian~DO Anderson.
\newblock Reverse-time diffusion equation models.
\newblock \emph{Stochastic Processes and their Applications}, 12\penalty0 (3):\penalty0 313--326, 1982.

\bibitem[Austin et~al.(2021)Austin, Johnson, Ho, Tarlow, and Van Den~Berg]{austin2021structured}
Jacob Austin, Daniel~D Johnson, Jonathan Ho, Daniel Tarlow, and Rianne Van Den~Berg.
\newblock Structured denoising diffusion models in discrete state-spaces.
\newblock \emph{Advances in neural information processing systems}, 34:\penalty0 17981--17993, 2021.

\bibitem[Bai and Urtasun(2017)]{bai2017deep}
Min Bai and Raquel Urtasun.
\newblock Deep watershed transform for instance segmentation.
\newblock In \emph{Proceedings of the IEEE conference on computer vision and pattern recognition}, pages 5221--5229, 2017.

\bibitem[Campbell et~al.(2022)Campbell, Benton, De~Bortoli, Rainforth, Deligiannidis, and Doucet]{campbell2022continuous}
Andrew Campbell, Joe Benton, Valentin De~Bortoli, Thomas Rainforth, George Deligiannidis, and Arnaud Doucet.
\newblock A continuous time framework for discrete denoising models.
\newblock \emph{Advances in Neural Information Processing Systems}, 35:\penalty0 28266--28279, 2022.

\bibitem[Chen et~al.(2017)Chen, Papandreou, Kokkinos, Murphy, and Yuille]{chen2017deeplab}
Liang-Chieh Chen, George Papandreou, Iasonas Kokkinos, Kevin Murphy, and Alan~L Yuille.
\newblock Deeplab: Semantic image segmentation with deep convolutional nets, atrous convolution, and fully connected crfs.
\newblock \emph{IEEE transactions on pattern analysis and machine intelligence}, 40\penalty0 (4):\penalty0 834--848, 2017.

\bibitem[Chen et~al.(2018)Chen, Zhu, Papandreou, Schroff, and Adam]{chen2018encoder}
Liang-Chieh Chen, Yukun Zhu, George Papandreou, Florian Schroff, and Hartwig Adam.
\newblock Encoder-decoder with atrous separable convolution for semantic image segmentation.
\newblock In \emph{Proceedings of the European conference on computer vision (ECCV)}, pages 801--818, 2018.

\bibitem[Cheng et~al.(2021)Cheng, Girshick, Doll{\'a}r, Berg, and Kirillov]{cheng2021boundary}
Bowen Cheng, Ross Girshick, Piotr Doll{\'a}r, Alexander~C Berg, and Alexander Kirillov.
\newblock Boundary iou: Improving object-centric image segmentation evaluation.
\newblock In \emph{Proceedings of the IEEE/CVF conference on computer vision and pattern recognition}, pages 15334--15342, 2021.

\bibitem[Cheng et~al.(2022)Cheng, Misra, Schwing, Kirillov, and Girdhar]{cheng2022masked}
Bowen Cheng, Ishan Misra, Alexander~G Schwing, Alexander Kirillov, and Rohit Girdhar.
\newblock Masked-attention mask transformer for universal image segmentation.
\newblock In \emph{Proceedings of the IEEE/CVF conference on computer vision and pattern recognition}, pages 1290--1299, 2022.

\bibitem[Cheng et~al.(2020)Cheng, Chung, Tai, and Tang]{cheng2020cascadepsp}
Ho~Kei Cheng, Jihoon Chung, Yu-Wing Tai, and Chi-Keung Tang.
\newblock Cascadepsp: Toward class-agnostic and very high-resolution segmentation via global and local refinement.
\newblock In \emph{Proceedings of the IEEE/CVF conference on computer vision and pattern recognition}, pages 8890--8899, 2020.

\bibitem[Cordts et~al.(2016)Cordts, Omran, Ramos, Rehfeld, Enzweiler, Benenson, Franke, Roth, and Schiele]{cordts2016cityscapes}
Marius Cordts, Mohamed Omran, Sebastian Ramos, Timo Rehfeld, Markus Enzweiler, Rodrigo Benenson, Uwe Franke, Stefan Roth, and Bernt Schiele.
\newblock The cityscapes dataset for semantic urban scene understanding.
\newblock In \emph{Proceedings of the IEEE conference on computer vision and pattern recognition}, pages 3213--3223, 2016.

\bibitem[Everingham et~al.(2010)Everingham, Van~Gool, Williams, Winn, and Zisserman]{everingham2010pascal}
Mark Everingham, Luc Van~Gool, Christopher~KI Williams, John Winn, and Andrew Zisserman.
\newblock The pascal visual object classes (voc) challenge.
\newblock \emph{International journal of computer vision}, 88\penalty0 (2):\penalty0 303--338, 2010.

\bibitem[Fang et~al.(2021)Fang, Yang, Wang, Li, Fang, Shan, Feng, and Liu]{fang2021instances}
Yuxin Fang, Shusheng Yang, Xinggang Wang, Yu~Li, Chen Fang, Ying Shan, Bin Feng, and Wenyu Liu.
\newblock Instances as queries.
\newblock In \emph{Proceedings of the IEEE/CVF international conference on computer vision}, pages 6910--6919, 2021.

\bibitem[Fu et~al.(2025)Fu, Lou, and Yu]{fu2025segman}
Yunxiang Fu, Meng Lou, and Yizhou Yu.
\newblock Segman: Omni-scale context modeling with state space models and local attention for semantic segmentation.
\newblock In \emph{2025 IEEE/CVF Conference on Computer Vision and Pattern Recognition (CVPR)}, pages 19077--19087. IEEE, 2025.

\bibitem[Guo et~al.(2022)Guo, Lu, Hou, Liu, Cheng, and Hu]{guo2022segnext}
Meng-Hao Guo, Cheng-Ze Lu, Qibin Hou, Zhengning Liu, Ming-Ming Cheng, and Shi-Min Hu.
\newblock Segnext: Rethinking convolutional attention design for semantic segmentation.
\newblock \emph{Advances in neural information processing systems}, 35:\penalty0 1140--1156, 2022.

\bibitem[Gupta et~al.(2019)Gupta, Dollar, and Girshick]{gupta2019lvis}
Agrim Gupta, Piotr Dollar, and Ross Girshick.
\newblock Lvis: A dataset for large vocabulary instance segmentation.
\newblock In \emph{Proceedings of the IEEE/CVF conference on computer vision and pattern recognition}, pages 5356--5364, 2019.

\bibitem[Hayder et~al.(2017)Hayder, He, and Salzmann]{hayder2017boundary}
Zeeshan Hayder, Xuming He, and Mathieu Salzmann.
\newblock Boundary-aware instance segmentation.
\newblock In \emph{Proceedings of the IEEE conference on computer vision and pattern recognition}, pages 5696--5704, 2017.

\bibitem[He et~al.(2017)He, Gkioxari, Doll{\'a}r, and Girshick]{he2017mask}
Kaiming He, Georgia Gkioxari, Piotr Doll{\'a}r, and Ross Girshick.
\newblock Mask r-cnn.
\newblock In \emph{Proceedings of the IEEE international conference on computer vision}, pages 2961--2969, 2017.

\bibitem[Ho et~al.(2020)Ho, Jain, and Abbeel]{ho2020denoising}
Jonathan Ho, Ajay Jain, and Pieter Abbeel.
\newblock Denoising diffusion probabilistic models.
\newblock \emph{Advances in neural information processing systems}, 33:\penalty0 6840--6851, 2020.

\bibitem[Ke et~al.(2022)Ke, Danelljan, Li, Tai, Tang, and Yu]{ke2022mask}
Lei Ke, Martin Danelljan, Xia Li, Yu-Wing Tai, Chi-Keung Tang, and Fisher Yu.
\newblock Mask transfiner for high-quality instance segmentation.
\newblock In \emph{Proceedings of the IEEE/CVF Conference on Computer Vision and Pattern Recognition}, pages 4412--4421, 2022.

\bibitem[Kirillov et~al.(2020)Kirillov, Wu, He, and Girshick]{kirillov2020pointrend}
Alexander Kirillov, Yuxin Wu, Kaiming He, and Ross Girshick.
\newblock Pointrend: Image segmentation as rendering.
\newblock In \emph{Proceedings of the IEEE/CVF conference on computer vision and pattern recognition}, pages 9799--9808, 2020.

\bibitem[Kr{\"a}henb{\"u}hl and Koltun(2011)]{krahenbuhl2011efficient}
Philipp Kr{\"a}henb{\"u}hl and Vladlen Koltun.
\newblock Efficient inference in fully connected crfs with gaussian edge potentials.
\newblock \emph{Advances in neural information processing systems}, 24, 2011.

\bibitem[Lin et~al.(2014)Lin, Maire, Belongie, Hays, Perona, Ramanan, Doll{\'a}r, and Zitnick]{lin2014microsoft}
Tsung-Yi Lin, Michael Maire, Serge Belongie, James Hays, Pietro Perona, Deva Ramanan, Piotr Doll{\'a}r, and C~Lawrence Zitnick.
\newblock Microsoft coco: Common objects in context.
\newblock In \emph{European conference on computer vision}, pages 740--755. Springer, 2014.

\bibitem[Liu et~al.(2025)Liu, Yu, Huang, Hua, Liu, Ma, Wang, Du, and Sun]{liu2025refinecatdiff}
Feng Liu, Hongjian Yu, Jiaan Huang, Xin Hua, Hengjia Liu, Zhimian Ma, Zhen Wang, Zhijiang Du, and Lining Sun.
\newblock Refinecatdiff: Toward high-quality medical image segmentation via a categorical diffusion refinement framework.
\newblock \emph{Advanced Intelligent Systems}, 7\penalty0 (12):\penalty0 e202401125, 2025.

\bibitem[Long et~al.(2015)Long, Shelhamer, and Darrell]{long2015fully}
Jonathan Long, Evan Shelhamer, and Trevor Darrell.
\newblock Fully convolutional networks for semantic segmentation.
\newblock In \emph{Proceedings of the IEEE conference on computer vision and pattern recognition}, pages 3431--3440, 2015.

\bibitem[Lou et~al.(2023)Lou, Meng, and Ermon]{lou2023discrete}
Aaron Lou, Chenlin Meng, and Stefano Ermon.
\newblock Discrete diffusion modeling by estimating the ratios of the data distribution.
\newblock \emph{arXiv preprint arXiv:2310.16834}, 2023.

\bibitem[Peng et~al.(2022)Peng, Liu, Tang, Hao, Chu, Chen, Wu, Chen, Yu, Du, et~al.]{peng2022pp}
Juncai Peng, Yi~Liu, Shiyu Tang, Yuying Hao, Lutao Chu, Guowei Chen, Zewu Wu, Zeyu Chen, Zhiliang Yu, Yuning Du, et~al.
\newblock Pp-liteseg: A superior real-time semantic segmentation model.
\newblock \emph{arXiv preprint arXiv:2204.02681}, 2022.

\bibitem[Perazzi et~al.(2016)Perazzi, Pont-Tuset, McWilliams, Van~Gool, Gross, and Sorkine-Hornung]{perazzi2016benchmark}
Federico Perazzi, Jordi Pont-Tuset, Brian McWilliams, Luc Van~Gool, Markus Gross, and Alexander Sorkine-Hornung.
\newblock A benchmark dataset and evaluation methodology for video object segmentation.
\newblock In \emph{Proceedings of the IEEE conference on computer vision and pattern recognition}, pages 724--732, 2016.

\bibitem[Qin et~al.(2022)Qin, Dai, Hu, Fan, Shao, and Van~Gool]{qin2022highly}
Xuebin Qin, Hang Dai, Xiaobin Hu, Deng-Ping Fan, Ling Shao, and Luc Van~Gool.
\newblock Highly accurate dichotomous image segmentation.
\newblock In \emph{European Conference on Computer Vision}, pages 38--56. Springer, 2022.

\bibitem[Ronneberger et~al.(2015)Ronneberger, Fischer, and Brox]{ronneberger2015u}
Olaf Ronneberger, Philipp Fischer, and Thomas Brox.
\newblock U-net: Convolutional networks for biomedical image segmentation.
\newblock In \emph{International Conference on Medical image computing and computer-assisted intervention}, pages 234--241. Springer, 2015.

\bibitem[Shen et~al.(2022)Shen, Zhang, Qi, Kuen, Xie, Wu, Lin, and Jia]{shen2022high}
Tiancheng Shen, Yuechen Zhang, Lu~Qi, Jason Kuen, Xingyu Xie, Jianlong Wu, Zhe Lin, and Jiaya Jia.
\newblock High quality segmentation for ultra high-resolution images.
\newblock In \emph{Proceedings of the IEEE/CVF conference on computer vision and pattern recognition}, pages 1310--1319, 2022.

\bibitem[Song and Ermon(2019)]{song2019generative}
Yang Song and Stefano Ermon.
\newblock Generative modeling by estimating gradients of the data distribution.
\newblock \emph{Advances in neural information processing systems}, 32, 2019.

\bibitem[Song et~al.(2020)Song, Sohl-Dickstein, Kingma, Kumar, Ermon, and Poole]{song2020score}
Yang Song, Jascha Sohl-Dickstein, Diederik~P Kingma, Abhishek Kumar, Stefano Ermon, and Ben Poole.
\newblock Score-based generative modeling through stochastic differential equations.
\newblock \emph{arXiv preprint arXiv:2011.13456}, 2020.

\bibitem[Sun et~al.(2022)Sun, Yu, Dai, Schuurmans, and Dai]{sun2022score}
Haoran Sun, Lijun Yu, Bo~Dai, Dale Schuurmans, and Hanjun Dai.
\newblock Score-based continuous-time discrete diffusion models.
\newblock \emph{arXiv preprint arXiv:2211.16750}, 2022.

\bibitem[Tang et~al.(2021)Tang, Chen, Li, Li, Zhang, and Hu]{tang2021look}
Chufeng Tang, Hang Chen, Xiao Li, Jianmin Li, Zhaoxiang Zhang, and Xiaolin Hu.
\newblock Look closer to segment better: Boundary patch refinement for instance segmentation.
\newblock In \emph{Proceedings of the IEEE/CVF conference on computer vision and pattern recognition}, pages 13926--13935, 2021.

\bibitem[Tian et~al.(2020)Tian, Shen, and Chen]{tian2020conditional}
Zhi Tian, Chunhua Shen, and Hao Chen.
\newblock Conditional convolutions for instance segmentation.
\newblock In \emph{European conference on computer vision}, pages 282--298. Springer, 2020.

\bibitem[Wang et~al.(2023)Wang, Ding, Liew, Liu, Zhao, and Wei]{wang2023segrefiner}
Mengyu Wang, Henghui Ding, Jun~Hao Liew, Jiajun Liu, Yao Zhao, and Yunchao Wei.
\newblock Segrefiner: Towards model-agnostic segmentation refinement with discrete diffusion process.
\newblock \emph{arXiv preprint arXiv:2312.12425}, 2023.

\bibitem[Wang et~al.(2020)Wang, Kong, Shen, Jiang, and Li]{wang2020solo}
Xinlong Wang, Tao Kong, Chunhua Shen, Yuning Jiang, and Lei Li.
\newblock Solo: Segmenting objects by locations.
\newblock In \emph{European conference on computer vision}, pages 649--665. Springer, 2020.

\bibitem[Wu et~al.(2025)Wu, Wang, Hong, Ji, Fu, Xu, Xu, and Jin]{wu2025medical}
Junde Wu, Ziyue Wang, Mingxuan Hong, Wei Ji, Huazhu Fu, Yanwu Xu, Min Xu, and Yueming Jin.
\newblock Medical sam adapter: Adapting segment anything model for medical image segmentation.
\newblock \emph{Medical image analysis}, 102:\penalty0 103547, 2025.

\bibitem[Xie et~al.(2021)Xie, Wang, Yu, Anandkumar, Alvarez, and Luo]{xie2021segformer}
Enze Xie, Wenhai Wang, Zhiding Yu, Anima Anandkumar, Jose~M Alvarez, and Ping Luo.
\newblock Segformer: Simple and efficient design for semantic segmentation with transformers.
\newblock \emph{Advances in neural information processing systems}, 34:\penalty0 12077--12090, 2021.

\bibitem[Yuan et~al.(2020)Yuan, Xie, Chen, and Wang]{yuan2020segfix}
Yuhui Yuan, Jingyi Xie, Xilin Chen, and Jingdong Wang.
\newblock Segfix: Model-agnostic boundary refinement for segmentation.
\newblock In \emph{European conference on computer vision}, pages 489--506. Springer, 2020.

\bibitem[Zhou et~al.(2020)Zhou, Price, Cohen, Wilensky, and Davis]{zhou2020deepstrip}
Peng Zhou, Brian Price, Scott Cohen, Gregg Wilensky, and Larry~S Davis.
\newblock Deepstrip: High-resolution boundary refinement.
\newblock In \emph{Proceedings of the IEEE/CVF conference on computer vision and pattern recognition}, pages 10558--10567, 2020.

\end{thebibliography}
\appendix

\section{Appendix}

\subsection{Multi-Class Refinement}
\label{subsec:multi_class}
\paragraph{Multi-step reverse sampling}
We additionally evaluate the full reverse diffusion process and include a strengthened D3PM variant, denoted as D3PM (modified), which adopts the same mixed training inputs as CDD, combining noisy masks and coarse masks during training.
Starting from a random noisy mask sampled from a uniform distribution, 
we iteratively apply each method's learned reverse transition from $t=T$ to $t=1$. 
At each step, each pixel is updated by selecting the state with the highest predicted reverse probability. 
The coarse mask is used only as conditioning input during sampling. 
This setting evaluates the model's ability to generate refined masks from noise via iterative denoising. Table~\ref{tab:appendix_d3pm_cdd_refined_epochs} shows that CDD provides more stable multi-step reverse refinement than D3PM, particularly at early training stages. 
At 50 epochs, D3PM exhibits less stable reverse sampling from noise, whereas CDD still produces better masks. 
As training progresses, both methods improve substantially at 100 epochs.
However, CDD consistently achieves higher IoU and BFScore, indicating more reliable performance during iterative reverse sampling. Experiments were conducted on NVIDIA RTX 4090 GPUs with 24GB memory.

\begin{figure*}[htbp]
  \centering
  \includegraphics[width=1.0\textwidth]{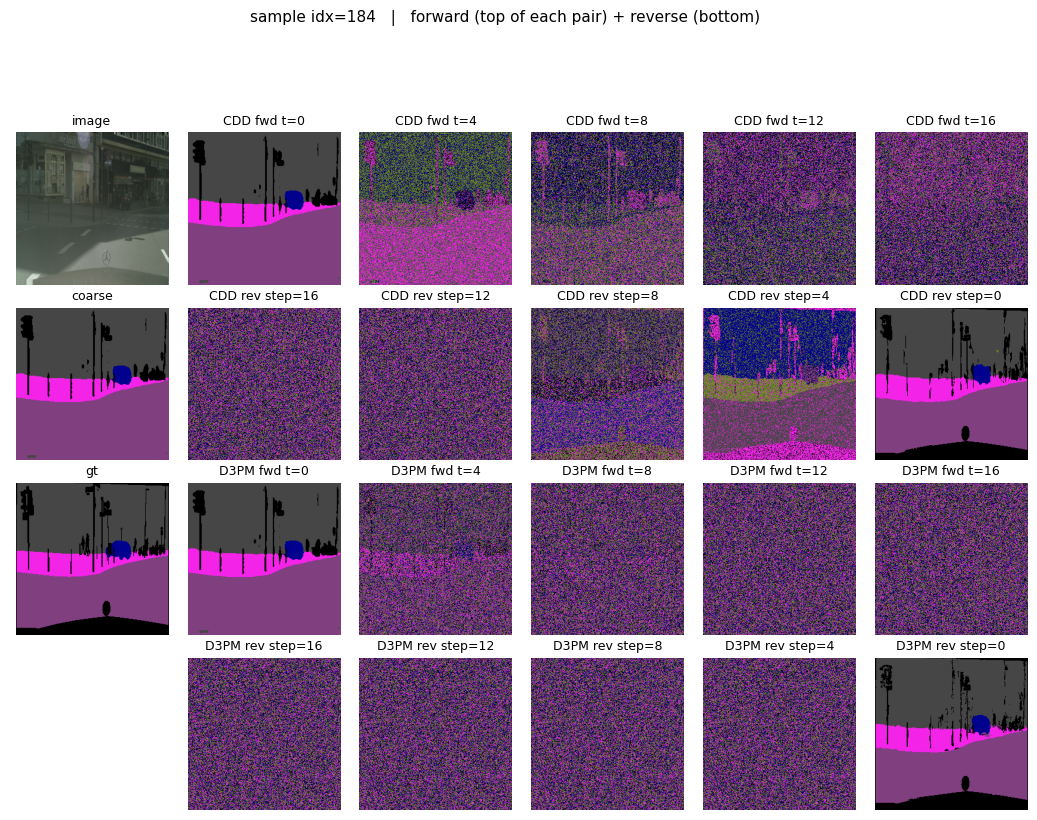}
    \caption{
    Visualization of forward and reverse dynamics for CDD and D3PM.
    The left column shows the reference images: input image, coarse mask, and ground-truth mask, while the fourth row is left empty.
    The first row shows the CDD forward process applied to the coarse mask at $t=0,4,8,12,16$.
    The second row shows snapshots of the CDD reverse sampling trajectory at steps $16,12,8,4,0$.
    The third row shows the D3PM forward corruption process at the same timesteps.
    The fourth row shows snapshots of the D3PM reverse sampling trajectory at the corresponding reverse steps. Both methods are visualized using identical timesteps and sampling schedules.
    }
  \label{fig:cdd_d3pm_dynamics}
\end{figure*}

We further visualize the forward and reverse diffusion trajectories of CDD and D3PM in Fig.~\ref{fig:cdd_d3pm_dynamics}. 
CDD corrupts masks through local cyclic transitions, which preserve partial spatial layouts and semantic structures along the forward trajectory. 
This structured degradation path makes the reverse process more interpretable, as semantic regions and boundaries are progressively recovered from noise. 
By contrast, D3PM rapidly approaches near-uniform noise, weakening the correspondence between intermediate states and the underlying semantic layout and leading to less structured reverse trajectories.
These observations suggest that CDD provides a structured local corruption process that preserves intermediate semantic structure, which is useful for segmentation refinement where spatial continuity and boundary preservation are essential.

\begin{table}[htbp]
\centering
\caption{Appendix comparison between D3PM and CDD under multi-step reverse refinement at different training epochs. 
Best results among refinement methods are in \textbf{bold}.}
\label{tab:appendix_d3pm_cdd_refined_epochs}
\small
\setlength{\tabcolsep}{6pt}
\begin{tabular}{lcccc}
\toprule
Method & Epochs & IoU & BFScore & $\Delta$IoU / $\Delta$BF \\
\midrule
Coarse mask & -- & 73.00 & 63.46 & -- / -- \\
\midrule
D3PM(modified) & 50  & 13.30 & 26.39 & -59.70 / -37.07 \\
CDD & 50  & \textbf{62.50} & \textbf{57.65} & \textbf{-10.50} / \textbf{-5.82} \\
\midrule
D3PM(modified)  & 100 & 81.78 & 77.52 & +8.78 / +14.06 \\
CDD  & 100 & \textbf{82.10} & \textbf{78.94} & \textbf{+9.10} / \textbf{+15.48}  \\
\bottomrule
\end{tabular}
\end{table}

\paragraph{Effect of label ordering.}
Since CDD defines a cyclic transition over the discrete label space, the order of semantic labels may influence the structure of the forward corruption process. 
We therefore test a less natural label order, where semantically related classes are separated in the cycle, For example, road is followed by vegetation rather than sidewalk. This makes the cyclic transition less semantically meaningful.

Table~\ref{tab:appendix_label_order} shows that CDD achieves nearly identical performance under the default and bad label orders. 
This suggests that although a semantically meaningful order provides a more interpretable transition structure, the learned refinement model is robust to moderate changes in label ordering.

\begin{table}[htbp]
\centering
\caption{Ablation on label ordering using Mask2Former coarse masks. 
Best results are in \textbf{bold}.}
\label{tab:appendix_label_order}
\small
\setlength{\tabcolsep}{6pt}
\begin{tabular}{lcccc}
\toprule
Label order & IoU & BFScore & $\Delta$IoU & $\Delta$BF \\
\midrule
Coarse mask & 73.00 & 63.46 & -- & -- \\
\midrule
Original order 
& \textbf{84.55} 
& 82.20 
& \textbf{+11.55} 
& +18.74 \\
Bad order 
& 84.52 
& \textbf{82.23} 
& +11.52 
& \textbf{+18.76} \\
\bottomrule
\end{tabular}
\end{table}

\paragraph{Per-class performance.}
Table~\ref{tab:per_class_iou} reports the per-class
IoU on Cityscapes. CDD improves all six categories,
with the largest gain observed in the heterogeneous
``other'' class.

\begin{table*}[t]
\centering
\caption{Per-class IoU (\%) on Cityscapes using
Mask2Former coarse masks.}
\label{tab:per_class_iou}
\small
\begin{tabular}{lccccccc}
\toprule
Method & mIoU & Road & Sidewalk & Building
& Vegetation & Car & Other \\
\midrule
Coarse & 73.00 & 80.96 & 71.42 & 76.09
& 83.14 & 87.37 & 39.01 \\
CDD & \textbf{84.55} & \textbf{93.07}
& \textbf{73.67} & \textbf{85.34}
& \textbf{88.45} & \textbf{91.52}
& \textbf{75.24} \\
\bottomrule
\end{tabular}
\end{table*}

\paragraph{Stability across random seeds.}
Across three independent training runs (seeds 0, 1, 2),
CDD achieves $84.52 \pm 0.10$ IoU and
$82.15 \pm 0.11$ BFScore, indicating
stable refinement performance.

\paragraph{Effect of training mixing ratio.}
We evaluate the proportion of real coarse masks used
during training, with the remaining samples generated
by the CDD forward process. As shown in
Table~\ref{tab:mixing_ratio}, training exclusively
on CDD noise severely degrades single-pass inference,
while training exclusively on coarse masks reduces
iterative refinement performance. The 50/50 strategy
achieves the best balance across both inference modes.

\begin{table}[htbp]
\centering
\caption{Effect of the training mixing ratio on
Cityscapes using Mask2Former coarse masks.
Results are reported as IoU/BFScore (\%).}
\label{tab:mixing_ratio}
\small
\setlength{\tabcolsep}{5pt}
\begin{tabular}{lcc}
\toprule
Real coarse ratio
& Single-pass
& Iterative ($T=16$) \\
\midrule
0\%   & 29.01 / 34.69 & 75.55 / 69.97 \\
50\%  & \textbf{84.55 / 82.20}
       & \textbf{84.23 / 81.85} \\
100\% & 84.53 / 82.18 & 81.31 / 76.60 \\
\bottomrule
\end{tabular}
\end{table}

\begin{figure*}[htbp]
  \centering
  \includegraphics[width=1.0\textwidth]{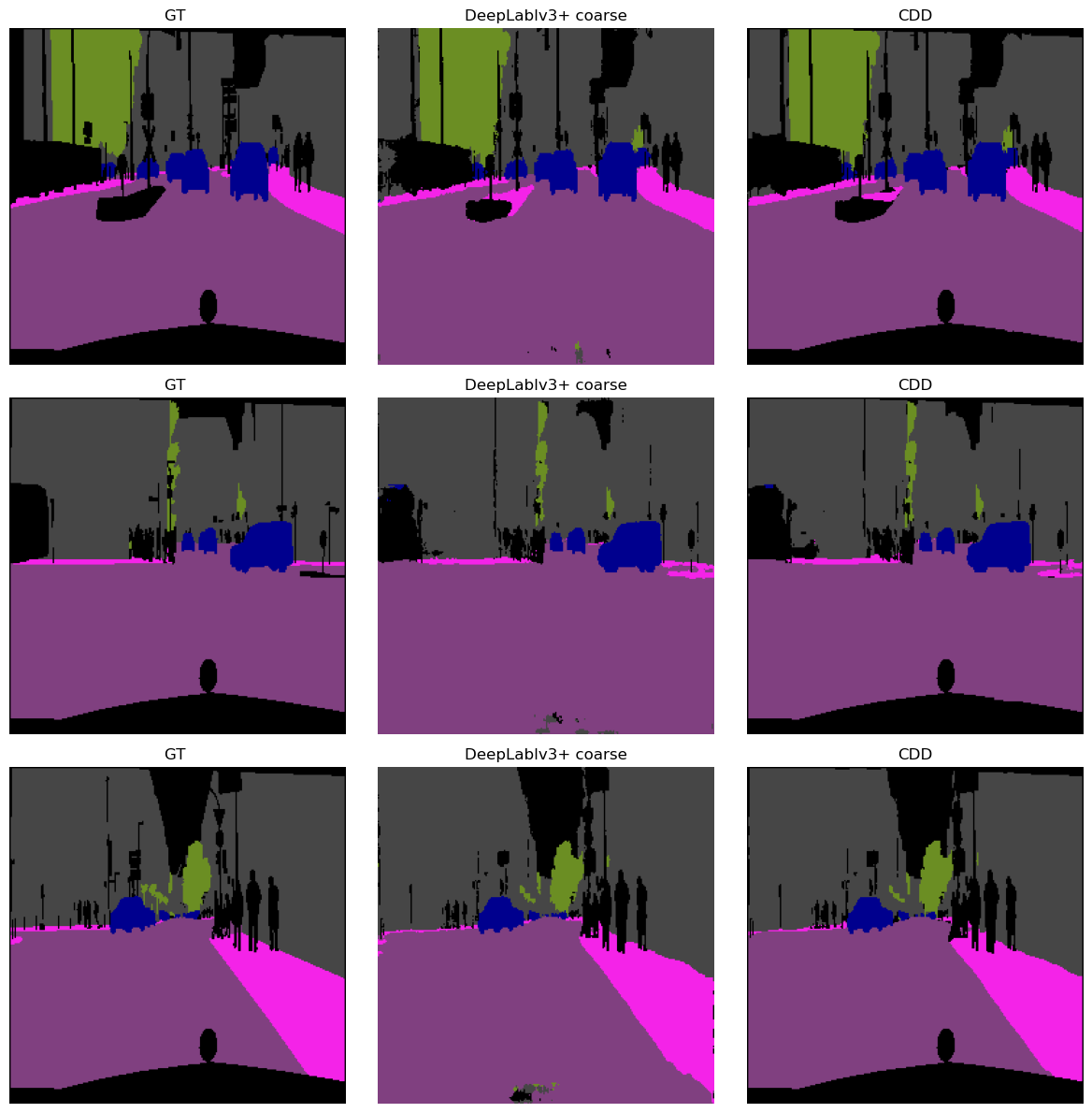}
    \caption{
Visualization of multi-class segmentation refinement on Cityscapes validation images.
Each row shows the ground-truth mask, the DeepLabv3+ coarse prediction, and the CDD-refined output.
CDD corrects local misclassifications and sharpens object boundaries while preserving the overall semantic structure of the coarse mask.
}
  \label{fig:cdd1}
\end{figure*}

\begin{figure*}[htbp]
  \centering
  \includegraphics[width=1.0\textwidth]{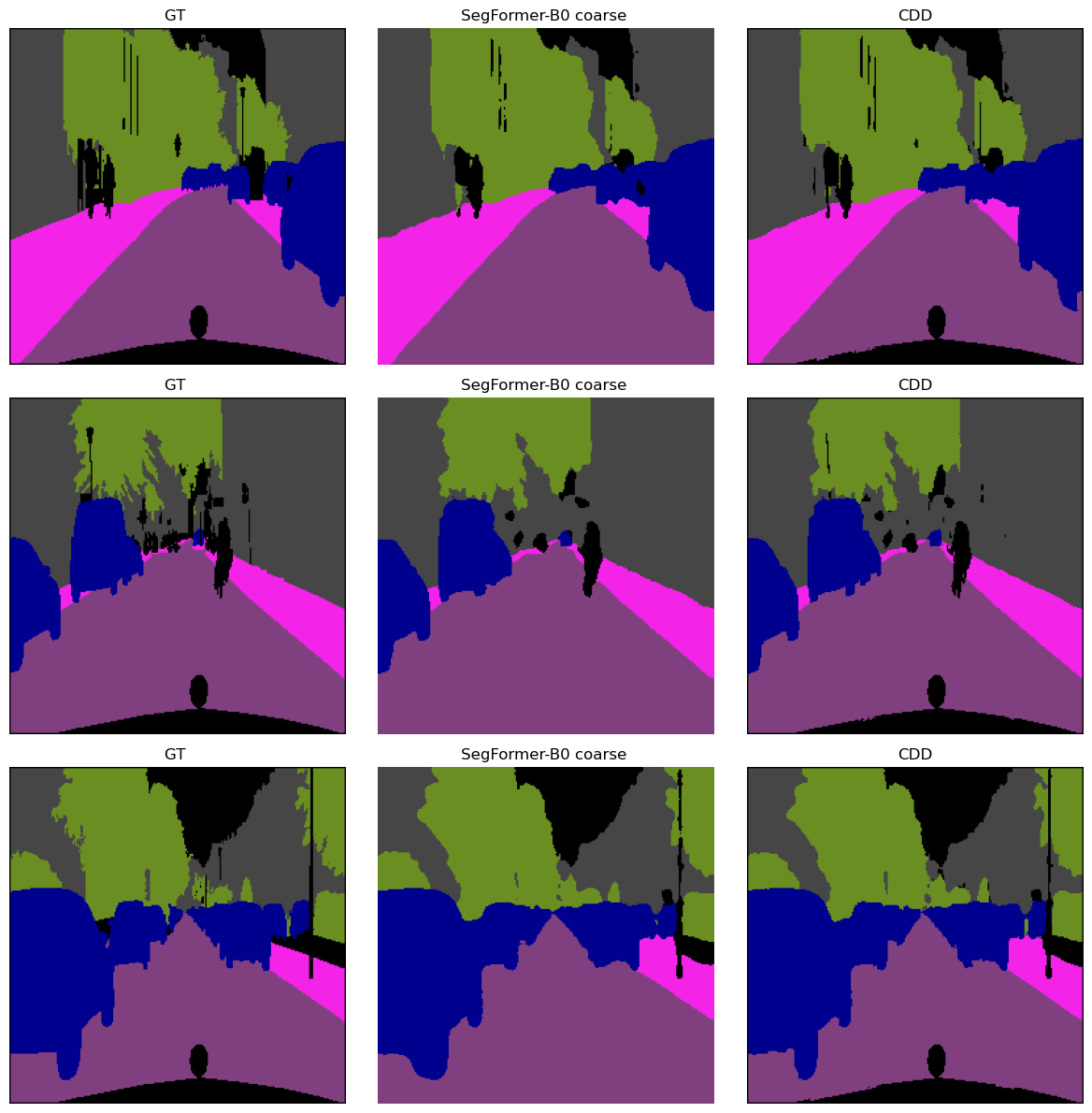}
    \caption{
Qualitative visualization of CDD refinement using SegFormer-B0 coarse masks on Cityscapes validation images.
From left to right, each row shows the ground-truth segmentation, the SegFormer-B0 coarse prediction, and the CDD-refined result.
    }
  \label{fig:cdd2}
\end{figure*}

\begin{figure*}[htbp]
  \centering
  \includegraphics[width=1.0\textwidth]{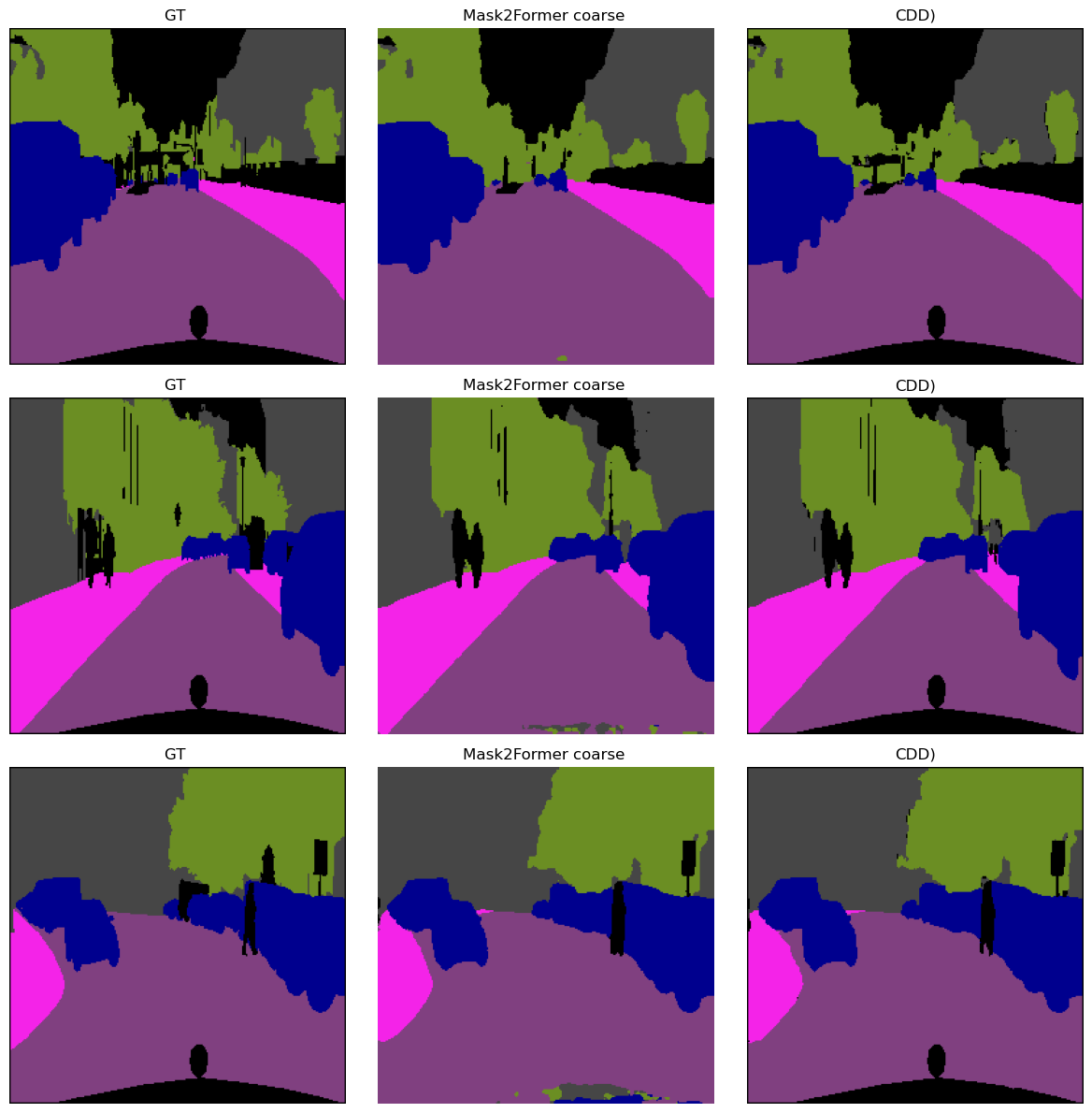}
  \caption{
Qualitative visualization of CDD refinement using Mask2Former coarse masks on Cityscapes validation images.
From left to right, each row shows the ground-truth segmentation, the Mask2Former coarse prediction, and the CDD-refined result.
}
  \label{fig:cdd3}
\end{figure*}

\subsection{Binary Refinement}
\paragraph{Implementation Details}
Our CDD refiner adopts the same UNet backbone and 
feature pyramid structure as SegRefiner~\cite{wang2023segrefiner}, 
ensuring that performance differences arise solely from the diffusion mechanism. 
For binary segmentation refinement, we set the number of diffusion steps to $T{=}6$ and the jump rate to $\lambda{=}0.3$, following the theoretical bound derived in Appendix~\ref{subsec:mix_rate}.

\subsubsection{Fine-grained Boundary Refinement on DIS5K}
\label{subsec:dis_result_section}
\begin{figure*}
    \centering        
    \includegraphics[width=1.0\textwidth]{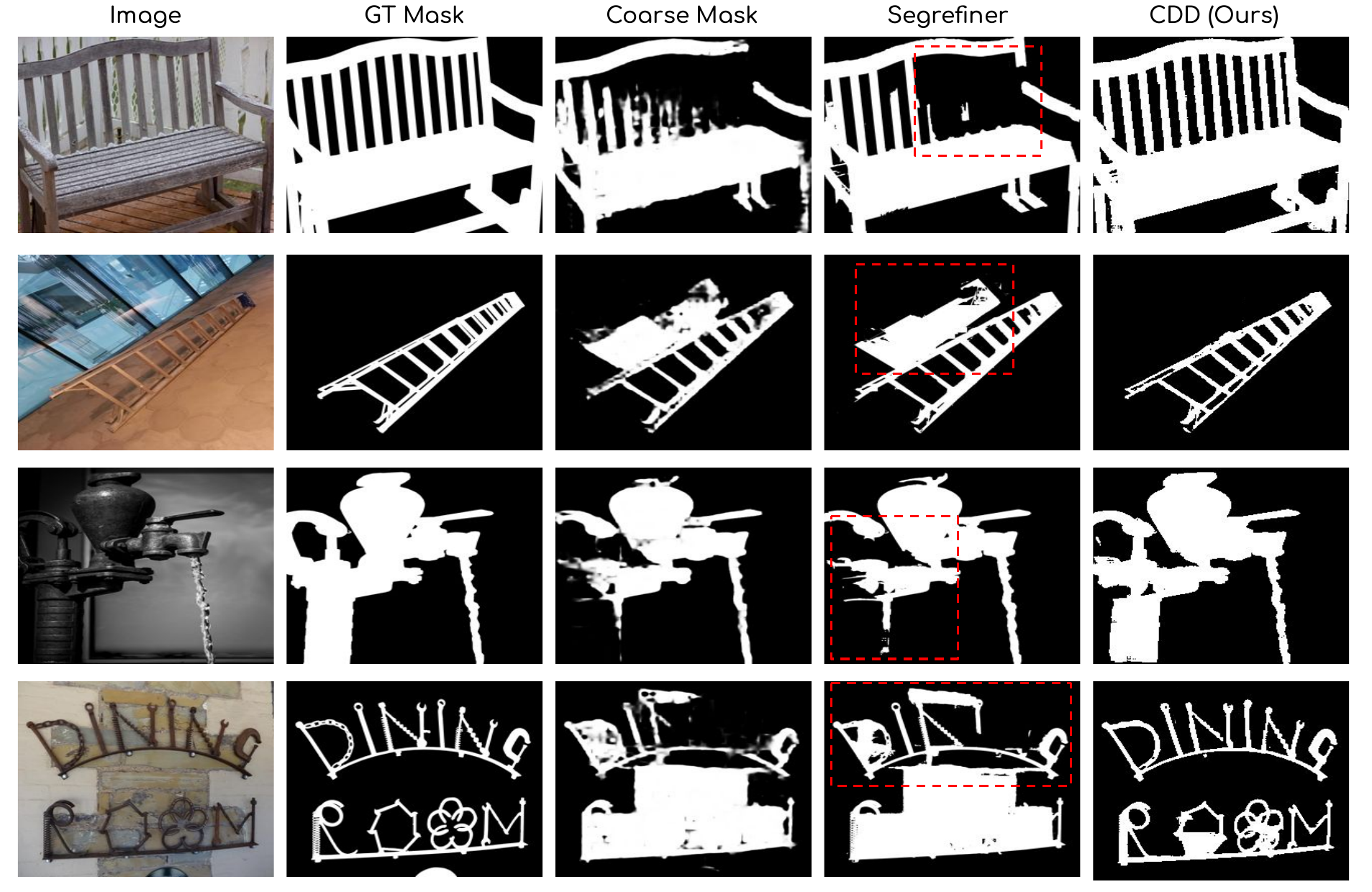}
    \caption{Qualitative comparison on the DIS dataset. 
    From left to right: input image, ground-truth mask, coarse mask, SegRefiner result, and our CDD refinement. 
    Compared with the coarse and SegRefiner predictions, CDD produces cleaner boundaries and more consistent fine structures, 
    especially for thin objects and intricate details such as ladder and text patterns.}
    \label{fig:dis_results}
\end{figure*}

We evaluate CDD on the high-resolution DIS5K benchmark to assess fine-grained boundary accuracy. DIS5K is a recently introduced dataset, and to our knowledge, 
SegRefiner~\cite{wang2023segrefiner} is the only method that has reported results on it so far.
The objective of this experiment is to assess how well our approach generalizes across different backbone architectures and to benchmark it fairly against SegRefiner.
As reported in Table~\ref{tab3:dis5k_refine}, CDD achieves higher IoU and mBA compared to segrefiner across multiple backbone architectures,
demonstrating superior capability in preserving fine object contours and boundary details. These results establish CDD as a strong and general model-agnostic refinement framework for both instance- and pixel-level segmentation tasks. Qualitative results in Fig.~\ref{fig:dis_results} further illustrate 
that CDD recovers fine object structures with sharper contours and cleaner shapes 
than SegRefiner, particularly on thin and intricate regions.
\begin{table*}[htbp]
\centering
\caption{Comparison between Ground Truth, SegRefiner~\cite{wang2023segrefiner}, and our CDD on DIS5K. 
Mean IoU (\%) and mean Boundary Accuracy (mBA, \%) are reported for the validation (VD) and four test sets (TE1–TE4). 
Green numbers indicate the improvement of Ours over SegRefiner.}
\label{tab3:dis5k_refine}
\small
\setlength{\tabcolsep}{3.9pt}
\renewcommand{\arraystretch}{1.15}
\resizebox{\linewidth}{!}{
\begin{tabular}{l c *{6}{ccc}}
\toprule
\multirow{2}{*}{Dataset} & \multirow{2}{*}{Metric} &
\multicolumn{3}{c}{U\text{-}Net} &
\multicolumn{3}{c}{PFNet} &
\multicolumn{3}{c}{PSPNet} &
\multicolumn{3}{c}{ICNet} &
\multicolumn{3}{c}{HRNet} &
\multicolumn{3}{c}{ISNet}\\
\cmidrule(lr){3-5}\cmidrule(lr){6-8}\cmidrule(lr){9-11}\cmidrule(lr){12-14}\cmidrule(lr){15-17}\cmidrule(lr){18-20}
& & GT & SegR & Ours & GT & SegR & Ours & GT & SegR & Ours & GT & SegR & Ours & GT & SegR & Ours & GT & SegR & Ours \\
\midrule
DIS-VD  & IoU  & 54.8 & 58.7 & \textbf{59.9}{\color{darkgreen}(+1.2)} & 56.2 & 60.4 & \textbf{61.3}{\color{darkgreen}(+0.9)} & 56.4 & 60.4 & \textbf{63.0}{\color{darkgreen}(+2.6)} & 56.5 & 61.4 & \textbf{61.7}{\color{darkgreen}(+0.3)} & 61.0 & 64.4 & \textbf{64.1}{\color{red}(-0.3)} & 67.1 & 68.3 & \textbf{67.3}{\color{red}(-1.0)} \\
         & mBA & 69.8 & 75.4 & \textbf{78.0}{\color{darkgreen}(+2.6)} & 63.7 & 74.8 & \textbf{77.6}{\color{darkgreen}(+2.8)} & 62.8 & 74.8 & \textbf{77.3}{\color{darkgreen}(+2.5)} & 66.5 & 75.6 & \textbf{77.9}{\color{darkgreen}(+2.3)} & 68.9 & 76.6 & \textbf{78.6}{\color{darkgreen}(+2.0)} & 74.1 & 76.6 & \textbf{78.4}{\color{darkgreen}(+1.8)} \\
\midrule
DIS-TE1 & IoU  & 44.1 & 47.1 & \textbf{51.9}{\color{darkgreen}(+4.8)} & 47.0 & 49.0 & \textbf{53.8}{\color{darkgreen}(+4.7)} & 47.6 & 49.0 & \textbf{55.3}{\color{darkgreen}(+6.3)} & 46.2 & 49.7 & \textbf{53.7}{\color{darkgreen}(+4.0)} & 51.0 & 53.0 & \textbf{55.6}{\color{darkgreen}(+2.7)} & 56.2 & 57.0 & \textbf{59.1}{\color{darkgreen}(+2.1)} \\
         & mBA & 70.1 & 75.0 & \textbf{78.8}{\color{darkgreen}(+3.8)} & 65.1 & 74.7 & \textbf{78.3}{\color{darkgreen}(+3.6)} & 64.4 & 74.7 & \textbf{78.2}{\color{darkgreen}(+3.6)} & 67.2 & 75.0 & \textbf{78.6}{\color{darkgreen}(+3.6)} & 69.3 & 75.6 & \textbf{78.9}{\color{darkgreen}(+3.3)} & 73.6 & 75.5 & \textbf{78.5}{\color{darkgreen}(+3.1)} \\
\midrule
DIS-TE2 & IoU  & 54.4 & 58.1 & \textbf{62.4}{\color{darkgreen}(+4.3)} & 57.0 & 60.3 & \textbf{64.8}{\color{darkgreen}(+4.6)} & 57.7 & 60.3 & \textbf{66.3}{\color{darkgreen}(+6.0)} & 56.5 & 60.3 & \textbf{64.1}{\color{darkgreen}(+3.8)} & 61.2 & 64.7 & \textbf{66.1}{\color{darkgreen}(+1.4)} & 67.0 & 67.8 & \textbf{68.6}{\color{darkgreen}(+0.7)} \\
         & mBA & 69.9 & 75.6 & \textbf{79.0}{\color{darkgreen}(+3.4)} & 64.6 & 75.1 & \textbf{78.6}{\color{darkgreen}(+3.6)} & 63.7 & 75.1 & \textbf{78.4}{\color{darkgreen}(+3.3)} & 66.8 & 75.6 & \textbf{78.8}{\color{darkgreen}(+3.2)} & 69.5 & 76.6 & \textbf{79.2}{\color{darkgreen}(+2.6)} & 74.0 & 76.4 & \textbf{78.0}{\color{darkgreen}(+1.6)} \\
\midrule
DIS-TE3 & IoU  & 59.3 & 63.3 & \textbf{66.1}{\color{darkgreen}(+2.8)} & 59.5 & 63.4 & \textbf{67.4}{\color{darkgreen}(+4.0)} & 59.9 & 63.4 & \textbf{67.8}{\color{darkgreen}(+4.4)} & 59.8 & 64.2 & \textbf{67.1}{\color{darkgreen}(+2.9)} & 64.4 & 67.6 & \textbf{69.2}{\color{darkgreen}(+1.6)} & 69.9 & 70.9 & \textbf{71.7}{\color{darkgreen}(+0.8)} \\
         & mBA & 70.6 & 76.7 & \textbf{79.3}{\color{darkgreen}(+2.6)} & 64.0 & 76.1 & \textbf{78.7}{\color{darkgreen}(+2.6)} & 62.7 & 76.1 & \textbf{78.5}{\color{darkgreen}(+2.4)} & 66.7 & 76.6 & \textbf{79.0}{\color{darkgreen}(+2.4)} & 69.5 & 77.6 & \textbf{79.6}{\color{darkgreen}(+2.0)} & 74.6 & 77.4 & \textbf{79.2}{\color{darkgreen}(+1.8)} \\
\midrule
DIS-TE4 & IoU  & 61.1 & 66.3 & \textbf{67.2}{\color{darkgreen}(+0.9)} & 59.1 & 66.6 & \textbf{67.7}{\color{darkgreen}(+1.1)} & 57.6 & 66.6 & \textbf{68.4}{\color{darkgreen}(+1.8)} & 60.5 & 67.0 & \textbf{68.0}{\color{darkgreen}(+1.0)} & 64.7 & 70.5 & \textbf{68.8}{\color{red}(-1.7)} & 70.1 & 72.2 & \textbf{70.9}{\color{red}(-1.3)} \\
         & mBA & 70.7 & 77.1 & \textbf{78.1}{\color{darkgreen}(+1.0)} & 62.8 & 76.4 & \textbf{77.6}{\color{darkgreen}(+1.2)} & 61.9 & 76.4 & \textbf{77.2}{\color{darkgreen}(+0.8)} & 66.4 & 77.1 & \textbf{77.9}{\color{darkgreen}(+0.8)} & 68.1 & 77.9 & \textbf{78.4}{\color{darkgreen}(+0.6)} & 74.3 & 77.7 & \textbf{78.1}{\color{darkgreen}(+0.5)} \\
\midrule
\textbf{Average} & IoU  & 54.9 & 58.7 & \textbf{61.5}{\color{darkgreen}(+2.8)} & 56.0 & 60.0 & \textbf{63.0}{\color{darkgreen}(+3.0)} & 56.1 & 60.0 & \textbf{64.1}{\color{darkgreen}(+4.1)} & 56.7 & 60.5 & \textbf{62.9}{\color{darkgreen}(+2.4)} & 61.1 & 64.0 & \textbf{64.8}{\color{darkgreen}(+0.7)} & 67.0 & 67.2 & \textbf{67.5}{\color{darkgreen}(+0.3)} \\
   & mBA & 70.2 & 76.0 & \textbf{78.6}{\color{darkgreen}(+2.6)} 
& 64.9 & 75.6 & \textbf{78.2}{\color{darkgreen}(+2.6)} 
& 64.1 & 75.6 & \textbf{77.9}{\color{darkgreen}(+2.3)} 
& 66.6 & 76.1 & \textbf{78.4}{\color{darkgreen}(+2.3)} 
& 69.3 & 76.9 & \textbf{78.9}{\color{darkgreen}(+2.0)} 
& 74.5 & 76.7 & \textbf{78.4}{\color{darkgreen}(+1.7)} \\
\bottomrule
\end{tabular}
}
\vspace{-1mm}
\end{table*}

\subsection{Effectiveness of KL Loss}
\paragraph{KL Loss (posterior matching).}
We train $f_\theta$ to match the model-induced reverse transition probabilities 
to the analytical posterior derived from the CDD forward process. 
Since the reverse transition at each pixel has only two admissible predecessors,
namely staying at the current label $x_t$ or moving one step backward to $x_t-1$ in the cyclic chain,
the model predicts a binary reverse distribution:
\[
p_\theta(x_{t-1}=x_t-1 \mid x_t,I,Y)=\sigma(z_\theta),
\qquad
p_\theta(x_{t-1}=x_t \mid x_t,I,Y)=1-\sigma(z_\theta).
\]
We then minimize
\[
\mathcal{L}_{\mathrm{KL}}
=
\mathbb{E}_{x_0,t,x_t}
\left[
D_{\mathrm{KL}}
\left(
q(x_{t-1}\mid x_t,x_0)
\,\|\, 
p_\theta(x_{t-1}\mid x_t,I,Y)
\right)
\right].
\]
Equivalently, this can be written as
\[
\begin{aligned}
\mathcal{L}_{\mathrm{KL}}
=
\mathbb{E}_{x_0,t,x_t}
\Big[
&
q(x_{t-1}=x_t-1\mid x_t,x_0)
\log
\frac{
q(x_{t-1}=x_t-1\mid x_t,x_0)
}{
p_\theta(x_{t-1}=x_t-1\mid x_t,I,Y)
}
\\
+
&
q(x_{t-1}=x_t\mid x_t,x_0)
\log
\frac{
q(x_{t-1}=x_t\mid x_t,x_0)
}{
p_\theta(x_{t-1}=x_t\mid x_t,I,Y)
}
\Big].
\end{aligned}
\]

For clarity, we distinguish between two training objectives:

\textbf{(1) Original binary segmentation objective.}
In all binary refinement experiments on COCO, LVIS, and DIS5K reported in the main paper,
we train $f_\theta$ to directly reconstruct the clean mask $x_0$ from an intermediate noisy state $x_t$
using the standard cross-entropy loss.
This setting is particularly effective for binary refinement.

\textbf{(2) KL Loss.}
In the following section (toy experiments), we additionally consider a KL-augmented objective
derived from the analytically known reverse transition of CDD.
This objective regularises the model to match the true stay-or-jump probabilities.
We empirically observe that this KL objective becomes more beneficial as the number of labels $K$ increases, 
highlighting its potential for multi-class refinement.

\paragraph{Ablation on Original objective vs.\ KL Loss}
We evaluate two training objectives for our discrete diffusion model: 
\begin{itemize}[leftmargin=*]
    \item \textbf{Original objective}: the standard cross-entropy loss for direct reconstruction.
    \item \textbf{Original objective+KL}: a KL-augmented objective that adds a  KL term ($0.001$) 
    to the CE loss. 
\end{itemize}
This ablation clarifies why we include the additional KL loss in our multi-class experiments and shows that posterior matching is important for general multi-class segmentation refinement.
For fairness, all ablations are performed under strictly controlled and reproducible conditions.  
We use a toy dataset of length $2000$ containing $64\times64$ masks composed of synthetic shapes, 
and corrupt them using our forward process with a fixed jump rate $\lambda = 0.3$.  
The time index $t$ for each corrupted sample is computed exactly using our analytical formula, 
ensuring that all models are trained under \textit{identical forward tolerances}.  
This fully controlled setup eliminates confounders such as semantics and appearance variability, 
allowing Original objective and Original objective+KL to be compared purely in terms of their learning behavior.  
For each value of $K$, we train Original objective and Original objective+KL side-by-side using the same U-Net architecture, 
optimizer, and hyperparameters. As $K$ increases, the reverse state space becomes more complex 
and optimization becomes more difficult.  

Results in Table~\ref{tab:kl_loss} show a clear trend.  
For small $K$, Original objective yields higher IoU, BFScore, and Boundary AP, confirming that direct 
supervision toward $x_0$ is a stronger learning signal for refinement, where the 
mapping from $x_t$ to $x_0$ is nearly deterministic.  
However, as $K$ grows, Original objective+KL becomes increasingly competitive and surpasses Original objective when $K \ge 4$, 
demonstrating that the KL component provides meaningful structural guidance when the 
reverse transition spans a richer discrete label space.  
In this paper, all our multi class refinement results are obtained using the Original objective+KL loss, since it provides the most stable and accurate supervision for multi class settings.
\begin{table}[h]
\centering
\small
\setlength{\tabcolsep}{5pt}
\begin{tabular}{c c l c c c}
\toprule
\textbf{K} & \textbf{T} & \textbf{Loss} 
& \textbf{IoU} $\uparrow$ 
& \textbf{BFScore} $\uparrow$ 
& \textbf{Boundary AP} $\uparrow$ \\
\midrule

\multirow{2}{*}{2} & \multirow{2}{*}{4} 
& \textbf{Original} 
& \textbf{0.9991} & \textbf{0.9915} & \textbf{0.9956} \\
& & Original+KL 
& 0.9989 & 0.9888 & 0.9943 \\

\midrule
\multirow{2}{*}{4} & \multirow{2}{*}{8} 
& Original 
& 0.9961 & 0.9766 & 0.9878 \\
& & \textbf{Original+KL} 
& \textbf{0.9962} & \textbf{0.9820} & \textbf{0.9907} \\

\midrule
\multirow{2}{*}{6} & \multirow{2}{*}{16} 
& Original 
& 0.9894 & \textbf{0.9701} & \textbf{0.9844} \\
& & Original+KL 
& \textbf{0.9905} & 0.9686 & 0.9834 \\

\midrule
\multirow{2}{*}{8} & \multirow{2}{*}{38} 
& Original 
& \textbf{0.9866} & 0.9489 & 0.9728 \\
& & Original+KL 
& 0.9852 & \textbf{0.9587} & \textbf{0.9782} \\

\midrule
\multirow{2}{*}{10} & \multirow{2}{*}{42} 
& Original 
& 0.9842 & 0.9461 & 0.9715 \\
& & \textbf{Original+KL} 
& \textbf{0.9855} & \textbf{0.9550} & \textbf{0.9762} \\

\bottomrule
\end{tabular}
\caption{Ablation of the KL posterior matching objective under different numbers of classes $K$. 
For each $K$, $T$ denotes the number of diffusion steps. 
Original+KL denotes the original objective combined with the KL posterior matching loss.}
\label{tab:kl_loss}
\end{table}

\subsection{Binary Segmentation Generation Example
}
\begin{figure*}[htbp]
  \centering
  \includegraphics[width=0.6\textwidth]{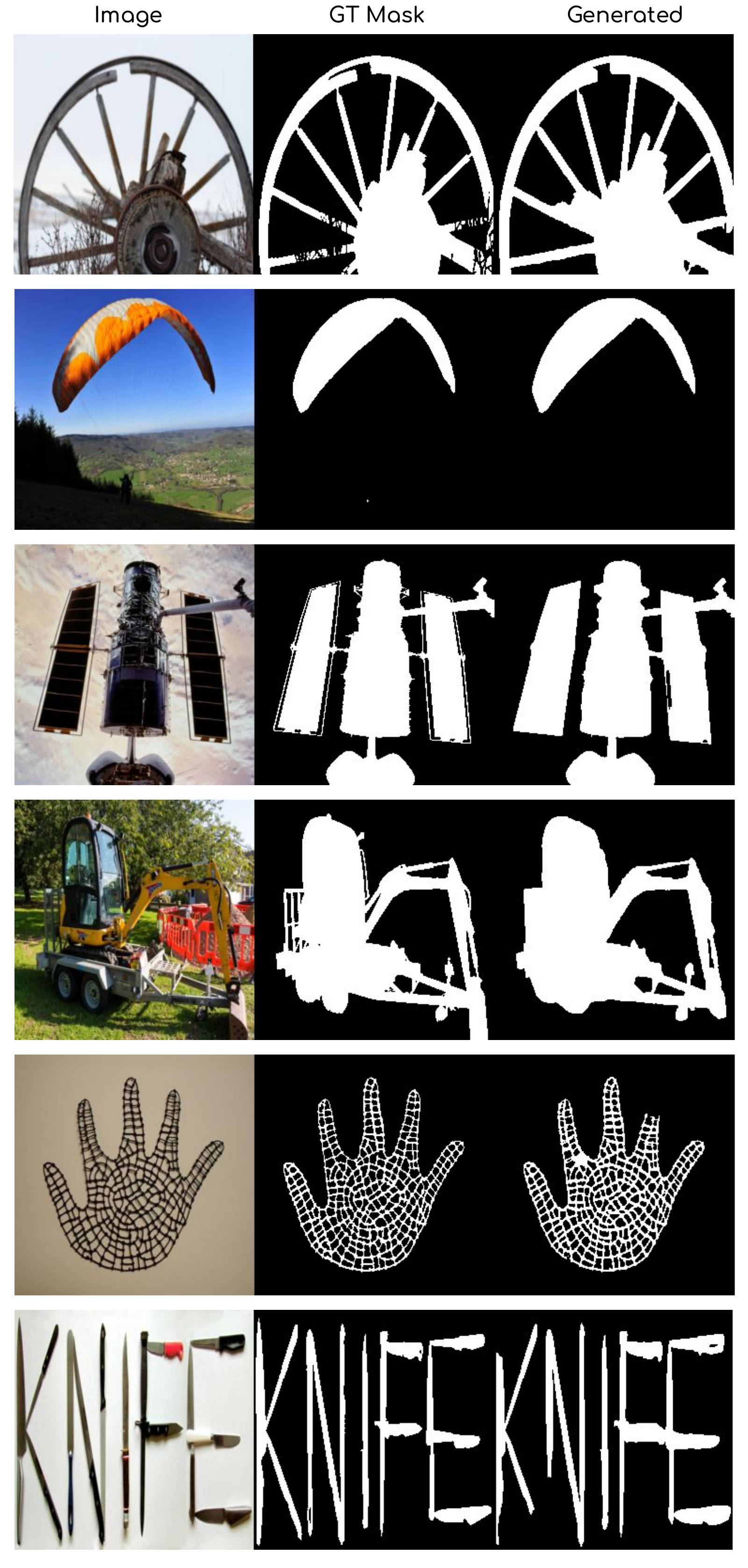}
  \caption{Results from our preliminary experiments on binary mask generation on the DIS5K dataset.}
  \label{fig:gen_results}
\end{figure*}

We also perform a preliminary experiment on unconditional binary mask generation
using the same model trained \emph{without} coarse-mask guidance on DIS.
This experiment is not part of our main contribution; it merely illustrates that the proposed CDD process can in principle be used as a standalone discrete generative model for segmentation masks. Quantitative results are reported in Table~\ref{tab:binary_generation}, and representative generations are shown in Figure~\ref{fig:gen_results}. As can be noted, the results in the unconditional case are worse as compared to the conditional results in the (Table 9). For instance, in VD, we obtained Mean IoU of 0.673 and mean mBA of 0.784 which is significantly better than these results in Table \ref{tab:binary_generation}.

\begin{table}[h]
\centering
\begin{tabular}{ccc}
\toprule
\textbf{Dataset Split} & \textbf{Mean IoU} & \textbf{Mean mBA} \\
\midrule
VD   & 0.5107 & 0.6201 \\
TE1  & 0.4677 & 0.6200 \\
TE2  & 0.5450 & 0.6299 \\
TE3  & 0.5692 & 0.6250 \\
\bottomrule
\end{tabular}
\caption{Binary Generation Results on DIS5K.}
\label{tab:binary_generation}
\end{table}


\subsection{Mixing rate}
\label{subsec:mix_rate}
To make the mixing behavior explicit, we write the forward process in matrix form.
For a single pixel, the Markov chain over the $K$ labels in $\mathcal{X}$ is governed by the $K\times K$ transition matrix $\Pi$, whose $(a,b)$-th entry is $q(x_{t+1}=b \mid x_t=a)$ from Eq.~(4).
Concretely,
\begin{equation}
\Pi =
\begin{bmatrix}
1-\lambda & \lambda     & 0           & \cdots      & 0 \\
0         & 1-\lambda   & \lambda     & \cdots      & 0 \\
\vdots    & \ddots      & \ddots      & \ddots      & \vdots \\
0         & \cdots      & 0           & 1-\lambda   & \lambda \\
\lambda   & 0           & \cdots      & 0           & 1-\lambda
\end{bmatrix},
\label{eq:transition-matrix}
\end{equation}
where each row corresponds to ``stay with probability $1-\lambda$'' and ``jump forward with probability $\lambda$'', and the last row wraps around to the first column due to the modulo-$K$ design.

This matrix is circulant: each row is a cyclic shift of the previous row.
As a result, its eigenvalues can be written in closed form as
\begin{equation}
\mu_m = (1-\lambda) + \lambda\, e^{2\pi i m / K}, \quad m = 0,1,\dots,K-1,
\label{eq:eigs}
\end{equation}
where $e^{2\pi i m / K}$ are the $K$-th roots of unity.
The largest eigenvalue is $\mu_0 = 1$, which corresponds to the stationary (uniform) state of the chain.
The remaining eigenvalues $\mu_m$ for $m\neq 0$ determine how fast the chain forgets its initial label and moves toward the stationary distribution.

The effective exponential decay rate is determined
by the largest modulus among the nontrivial
eigenvalues. From \eqref{eq:eigs}, we obtain
\begin{equation}
\gamma
= -\log\left(\max_{m\ne 0}|\mu_m|\right)
= -\frac{1}{2}\log\left[
1-2\lambda(1-\lambda)
\left(1-\cos\frac{2\pi}{K}\right)
\right].
\label{eq:spectral_gap_appendix}
\end{equation}
Indeed, the squared modulus of each eigenvalue satisfies
\begin{equation}
|\mu_m|^2
=
1-2\lambda(1-\lambda)
\left(1-\cos\frac{2\pi m}{K}\right).
\end{equation}

Here, $\gamma$ measures how quickly the chain mixes: a larger $\gamma$ implies faster convergence toward the uniform distribution.

This decay rate directly controls the decay of the distance between the current marginal distribution $\pi_t$ at time $t$ and the uniform distribution $u$.
In particular,
\begin{equation}
\|\pi_t - u\|_{\mathrm{TV}} \;\lesssim\; C\, e^{-\gamma t}, 
\qquad
t_{\mathrm{mix}}(\epsilon) \;\approx\; \frac{1}{\gamma}\,\log \frac{1}{\epsilon},
\label{eq:mixing-time}
\end{equation}
where $t_{\mathrm{mix}}(\epsilon)$ is the number of forward diffusion steps required for the chain to become $\epsilon$-close to uniform in total variation distance.
In practice, this means that the proposed cyclic jump process reaches an approximately uniform (high-entropy) state after a finite and predictable number of jumps, rather than requiring an unbounded number of diffusion steps.

For illustration, consider $K=5$ and $\lambda=0.3$.
The dominant nontrivial eigenvalues of the transition
matrix form a complex conjugate pair, with
\[
\mu_1=0.7+0.3e^{2\pi i/5}
\approx 0.7927+0.2853i.
\]
This gives an effective exponential decay rate
of $\gamma\approx 0.1714$.
Direct computation of the transition matrix powers,
starting from a deterministic label, gives
$t_{\mathrm{mix}}(0.1)=11$ and
$t_{\mathrm{mix}}(0.01)=25$.
These results confirm geometric convergence
toward the uniform stationary distribution.

\subsubsection{Impact of Jump Probability}
The jump probability $\lambda$ directly controls the corruption strength of the forward
process. A small $\lambda$ produces mild perturbations, where most labels remain unchanged
and the refinement problem becomes almost deterministic. Conversely, a large $\lambda$
aggressively mixes the label space, quickly destroying boundary structure and making the
reverse prediction substantially harder. 
From a spectral perspective, the mixing rate depends on
$\lambda(1-\lambda)$ rather than increasing monotonically
with $\lambda$. Within our tested range, excessively
small values lead to under-corruption, while larger
values promote faster mixing and may lead to
overly corrupted masks.

As shown in Table~\ref{tab:ablation_lambda}, this trade-off is reflected in performance:
$\lambda=0.1$ preserves structure too strongly, yielding slightly higher IoU but weaker
boundary accuracy; $\lambda=0.5$ over-corrupts the masks and degrades both metrics. 
The best overall performance is achieved at $\lambda=0.3$, which provides a balanced
corruption level that preserves global shape while injecting sufficient boundary noise for
effective refinement on DIS5K.

\begin{table}[h]
\centering
\caption{Ablation on the jump rate $\lambda$ (DIS validation set, $T=6$).}
\label{tab:ablation_lambda}
\begin{tabular}{lcc}
\toprule
\textbf{Setting} & \textbf{Mean IoU $\uparrow$} & \textbf{Mean mBA $\uparrow$} \\
\midrule
$T{=}6,\,\lambda{=}0.1$ & 0.5847 & 0.7702 \\
$T{=}6,\,\lambda{=}0.2$ & 0.5779 & 0.7716 \\
$T{=}6,\,\lambda{=}0.3$ & \textbf{0.5809} & \textbf{0.7732} \\
$T{=}6,\,\lambda{=}0.4$ & 0.5791 & 0.7731 \\
$T{=}6,\,\lambda{=}0.5$ & 0.5736 & 0.7668 \\
\bottomrule
\end{tabular}
\end{table}

\begin{figure*}[htbp]
  \centering
  \includegraphics[width=0.9\textwidth]{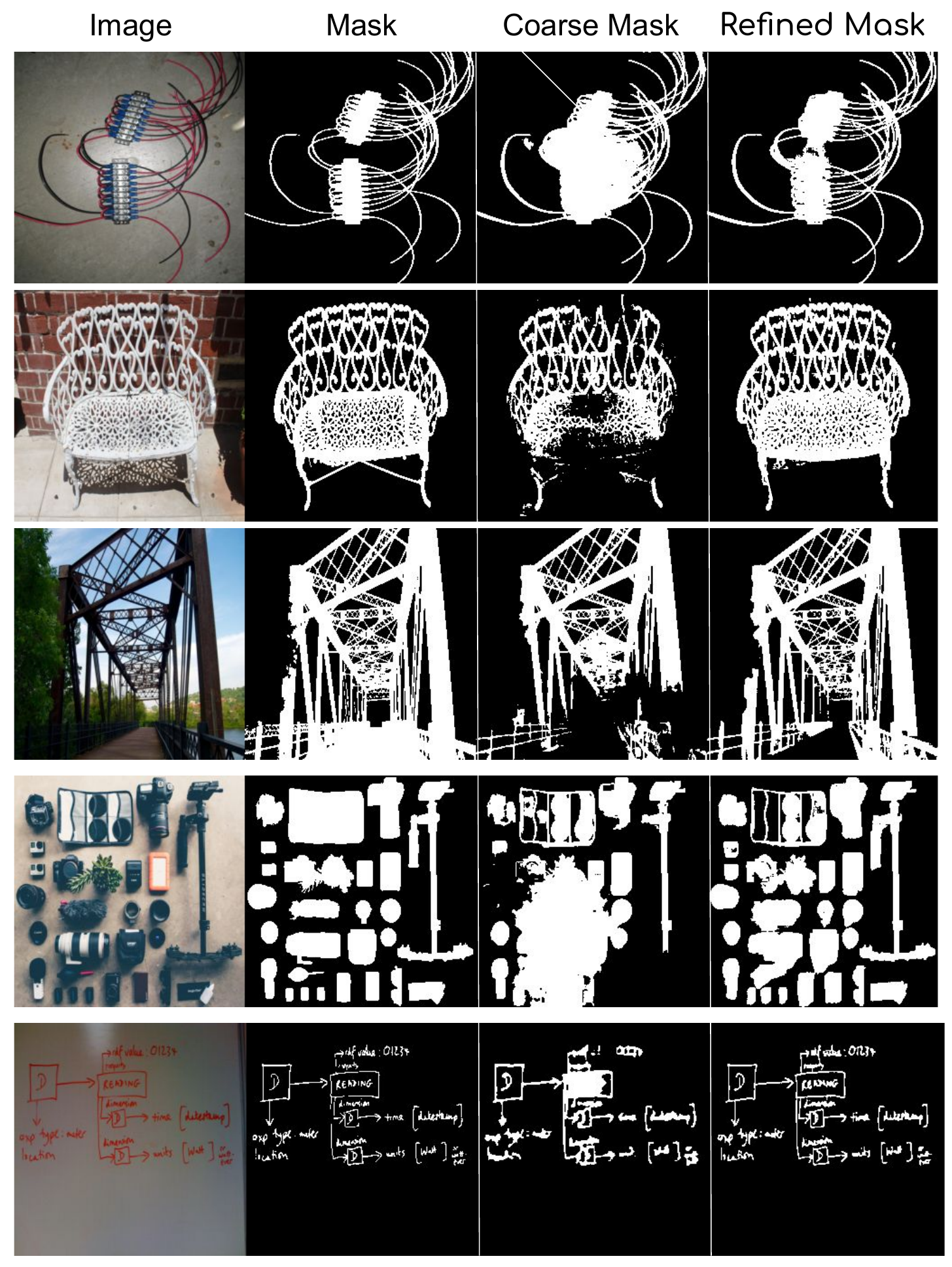}
  \caption{Refinement results on DIS dataset. From left to right: input image, ground-truth mask, coarse mask, and our refined mask. The coarse masks are taken from the UNet model and refined using our method. Notably, our method selectively corrects only the tampered regions while preserving the fine structures already present in the coarse mask.}
  \label{fig:refinement_dis}
\end{figure*}

\begin{figure*}[htbp]
  \centering
  \includegraphics[width=0.9\textwidth]{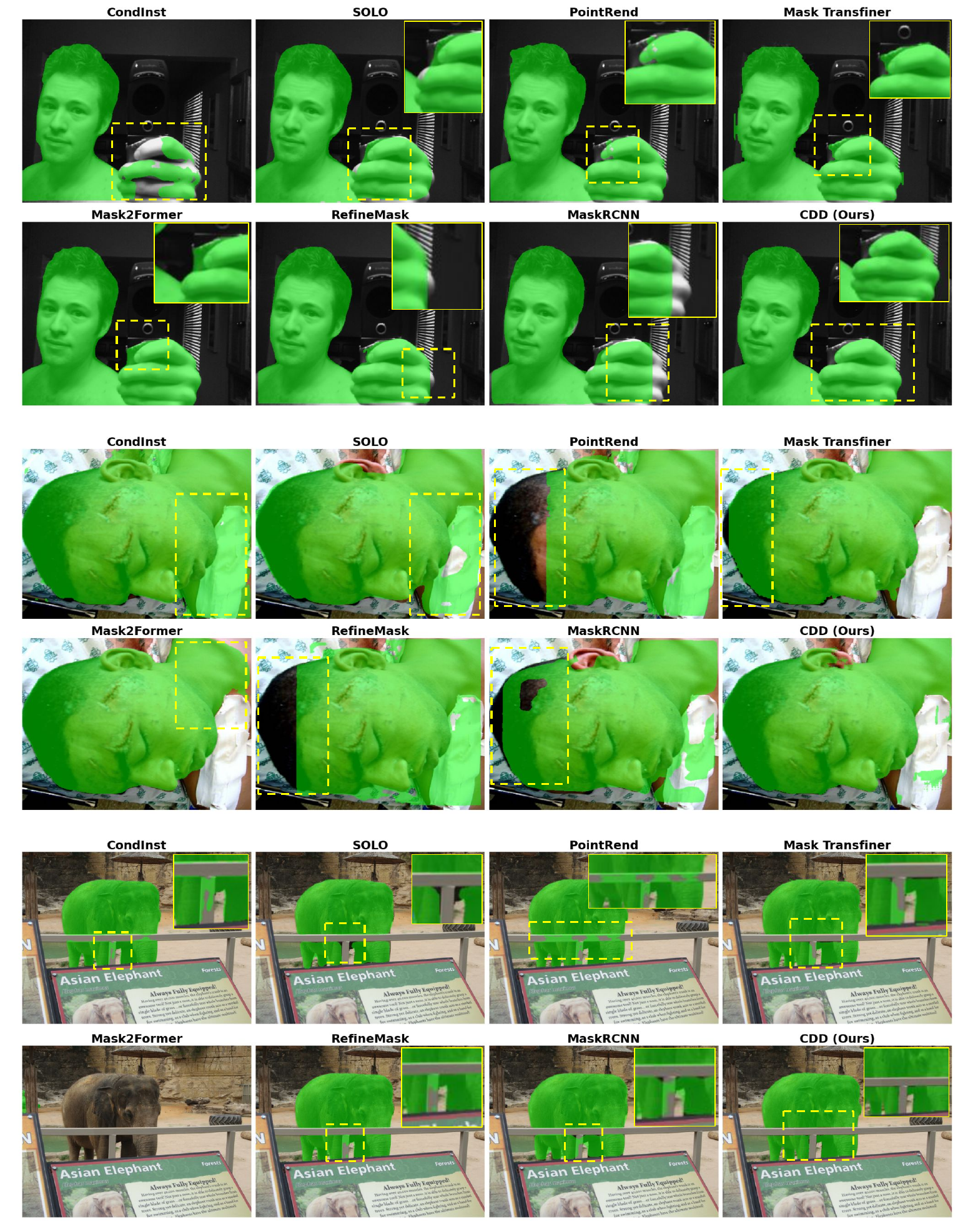}
  \caption{Refinement results on the COCO validation set. Instance segementation results are shown from 7 different mask generation models which are further refined by our method to produce the refined masks.}
  \label{fig:refinement_coco}
\end{figure*}

\section{Qualitative Analysis}
We provide further qualitative analysis for our method and provide more comparisons. We do that for the refinement on DIS dataset in fig. \ref{fig:refinement_dis}. As can be seen our refined mask substantially improves the coarse mask obtained from UNet model \cite{ronneberger2015u}. Particularly the thin wires in the first row, the wired seat of the chair in the second row, wired strussses in the bridge, the collection of objects in the fourth row and the equation on the board are all better segmented. However, challenges still remain on these very challenging cases. For instance, the road part of the bridge is not yet improved in the refined mask. 

In fig. \ref{fig:refinement_coco}, we consider the comparison with standard segmentation methods. In these examples we adopt the refinement of SOLO \cite{wang2020solo}. However, we observe similar refinement performance for the refinement of any of these standard segmentation methods. Particularly, we have also provided results with respect to MaskRCNN in table 3 in the main paper. As can be observed from the set of examples, we do a better job for challenging cases like the fingers in the first case, facial contours and hair in the second case, and the object edges in the third case.

\begin{figure*}[htbp]
  \centering
  \includegraphics[width=1.0\textwidth]{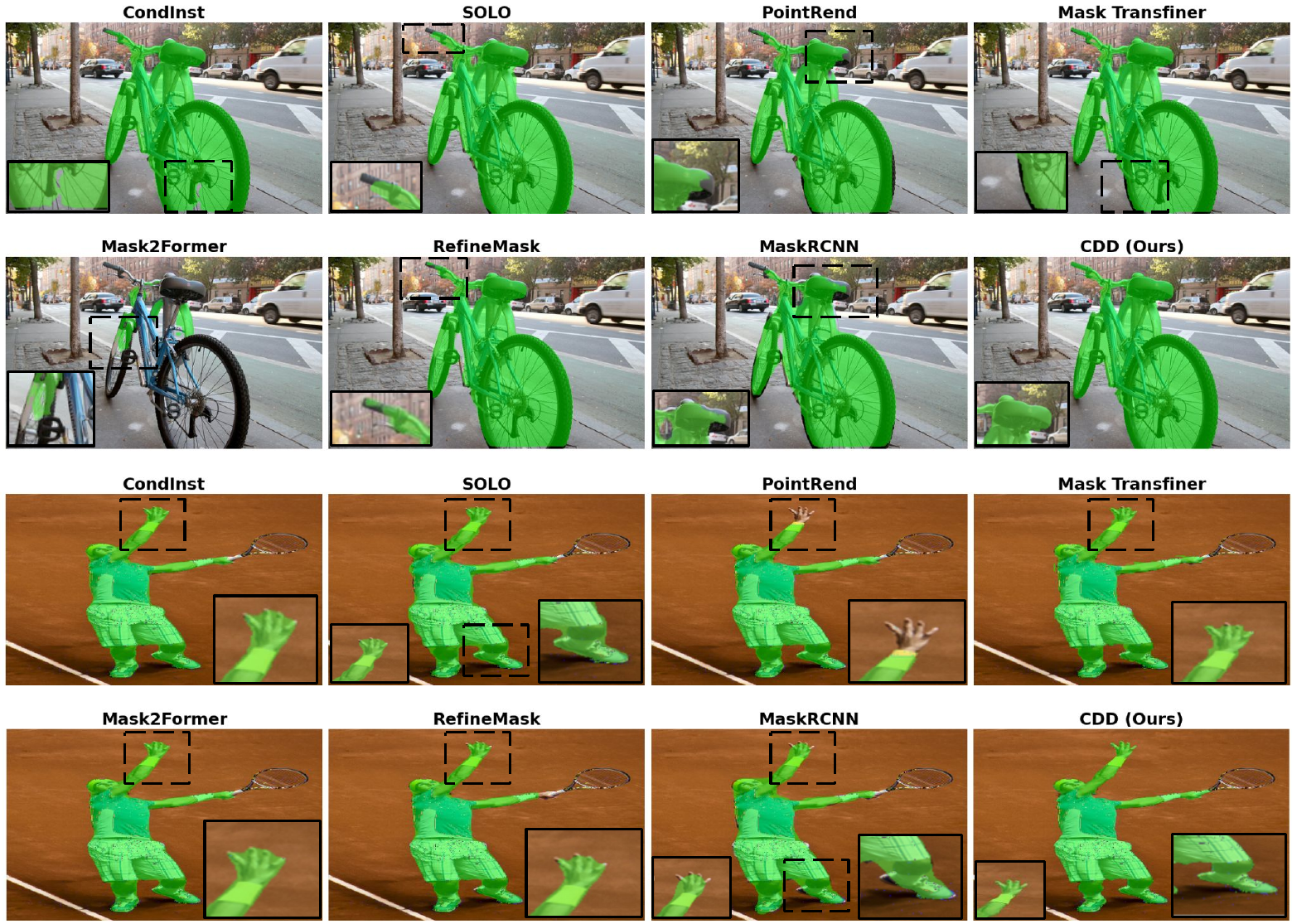}
  \caption{Visual comparison of refinement results on COCO corresponding to Table~\ref{tab2:instance_segmentation_models}. We provide qualitative comparisons with strong baseline instance segmentation methods. We can observe that the proposed method is more precise in comparison to all these baselines.
    (Best viewed with zoom.)}
  \label{fig:coco_lvis}
\end{figure*}

\end{document}